\documentclass[10pt,twocolumn,letterpaper]{article}

\usepackage{iccv}
\usepackage{times}
\usepackage{epsfig}
\usepackage{graphicx}
\usepackage{amsmath}
\usepackage{amssymb}
\usepackage{amsthm}
\usepackage{mathrsfs}
\usepackage{xspace}

\usepackage{multirow}
\usepackage{booktabs}
\usepackage{float}
\usepackage{threeparttable}
\usepackage{bm}
\usepackage{adjustbox}
\usepackage{array}
\usepackage{placeins}
\usepackage{afterpage}
\usepackage{stfloats}
\usepackage{textcomp}
\usepackage{bbding}

\usepackage{cite}

\usepackage{color, soul}
\sethlcolor{blue}

\newcommand*\rot{\multicolumn{1}{R{45}{1em}}}
\newcolumntype{R}[2]{%
    >{\adjustbox{angle=#1,lap=\width-(#2)}\bgroup}%
    l%
    <{\egroup}%
}

\usepackage[pagebackref=true,breaklinks=true,colorlinks,bookmarks=false]{hyperref}

\iccvfinalcopy

\def\iccvPaperID{3373}

\ificcvfinal\fi

\begin{document}

\title{DiffBEV: Conditional Diffusion Model for Bird's Eye View Perception}

\author{
Jiayu Zou\textsuperscript{\rm 1,3}\footnotemark[1]~~Zheng Zhu\textsuperscript{\rm 2}\footnotemark[2]~~Yun Ye\textsuperscript{\rm 2}\footnotemark[4]~~Xingang Wang\textsuperscript{\rm 1}\footnotemark[2]\\
\textsuperscript{\rm 1}Institute of Automation, Chinese Academy of Sciences
~ ~ \textsuperscript{\rm 2}PhiGent Robotics \\
~ ~ \textsuperscript{\rm 3}University of Chinese Academy of Sciences
}

\maketitle
\renewcommand{\thefootnote}{\fnsymbol{footnote}}
\footnotetext[1]{zoujiayu2020@ia.ac.cn}
\footnotetext[2]{Corresponding authors. zhengzhu@ieee.org, xingang.wang@ia.ac.cn}
\footnotetext[3]{\url{https://github.com/JiayuZou2020/DiffBEV}}
\footnotetext[4]{yun.ye@phigent.ai}
\ificcvfinal\thispagestyle{empty}\fi
\begin{abstract}
    BEV perception is of great importance in the field of autonomous driving, serving as the cornerstone of planning, controlling, and motion prediction. The quality of the BEV feature highly affects the performance of BEV perception. However, taking the noises in camera parameters and LiDAR scans into consideration, we usually obtain BEV representation with harmful noises. Diffusion models naturally have the ability to denoise noisy samples to the ideal data, which motivates us to utilize the diffusion model to get a better BEV representation. In this work, we propose an end-to-end framework, named DiffBEV, to exploit the potential of diffusion model to generate a more comprehensive BEV representation. To the best of our knowledge, we are the first to apply diffusion model to BEV perception. In practice, we design three types of conditions to guide the training of the diffusion model which denoises the coarse samples and refines the semantic feature in a progressive way. What's more, a cross-attention module is leveraged to fuse the context of BEV feature and the semantic content of conditional diffusion model. DiffBEV achieves a 25.9\% mIoU on the nuScenes dataset, which is 6.2\% higher than the best-performing existing approach. Quantitative and qualitative results on multiple benchmarks demonstrate the effectiveness of DiffBEV in BEV semantic segmentation and 3D object detection tasks. The code \footnotemark[3] will be available soon.
\end{abstract}

\section{Introduction}

\begin{figure}
    \centering
    \includegraphics[width=1.0\linewidth]{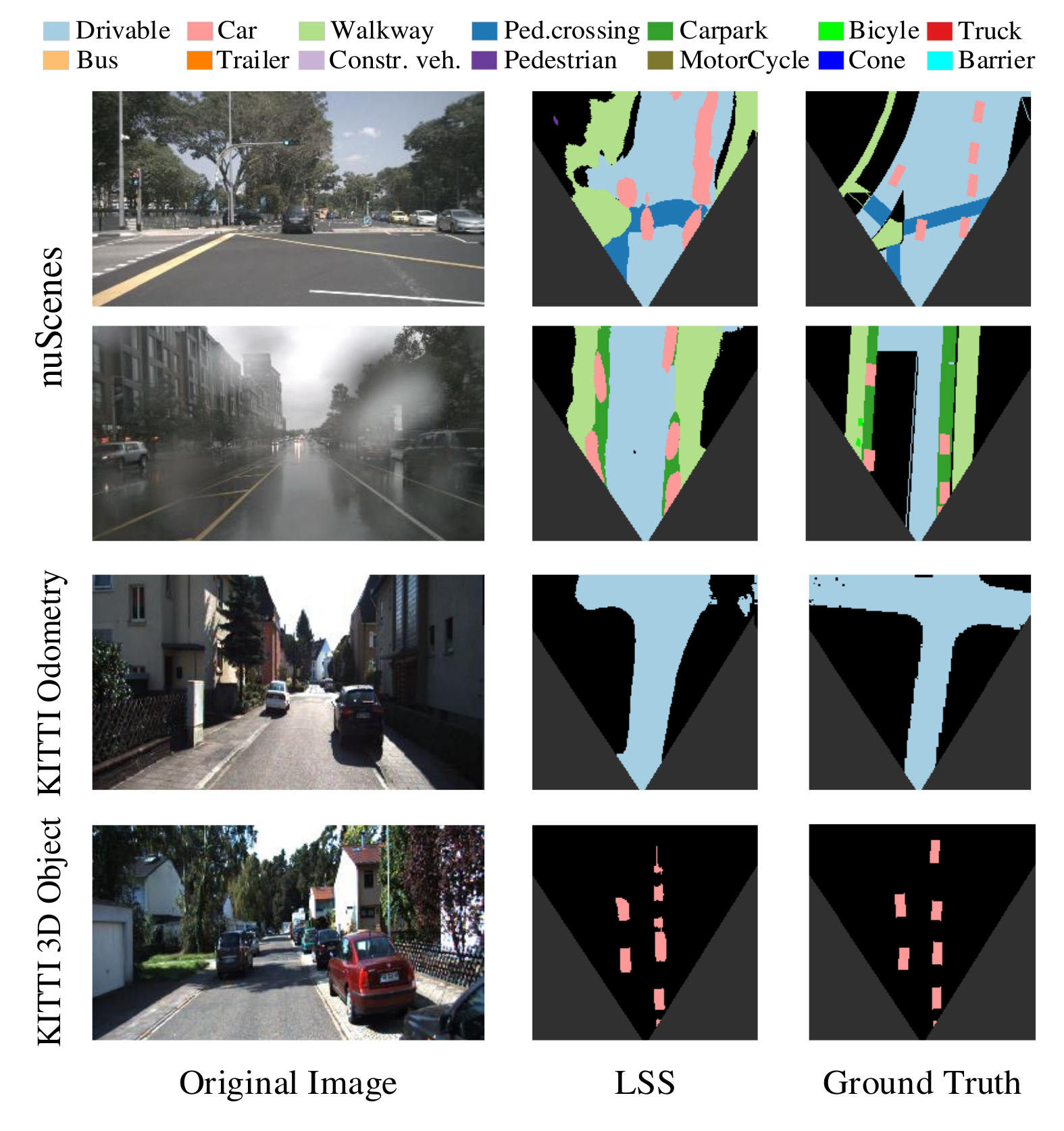}
    \caption{The poor segmentation results of LSS \cite{philion2020lift} model on the nuScenes and KITTI datasets.}
    \label{fig:intro}
\end{figure}

Bird's Eye View (BEV) perception plays a crucial role in autonomous driving tasks, which need a compact and accurate representation of the real world. One of the most important components of BEV perception is the quality of the BEV feature. Taking the classical LSS \cite{philion2020lift} as an illustration, it first extracts image features from the backbone encoder and then transforms them into BEV space along with depth estimation. However, the downstream perception results are often distorted, since the flat-world assumption is not always valid and the feature distribution in BEV is usually sparse.  As shown in Fig. \ref{fig:intro}, when LSS \cite{philion2020lift} is utilized as the view transformer, the final segmentation results have three deficiencies: (1) The prediction of dynamic object boundaries is ambiguous, where pixels of different vehicles are connected; (2) The perception of static areas such as the pedestrian crossing and walkway is too rough. In particular, there are a lot of redundant predictions on the nuScenes benchmark; (3) LSS \cite{philion2020lift} has a poor discriminative ability for background and foreground pixels. In the last two rows of Fig. \ref{fig:intro}, the interested drivable area and vehicle objects are misclassified into the background.

The above observations intuitively motivate us to explore more fine-grained and highly detailed BEV feature for downstream perception tasks. Taking the noises in camera parameters and LiDAR scans into consideration, we usually obtain BEV representation with harmful noises. Diffusion models naturally have the ability to denoise noisy samples to the ideal data. Recently, the diffusion probability models (DPM) have illustrated their great power in generative tasks \cite{diff-edit, diff-text-1,diff-inpaint,diff-motion-1}, but their potential in BEV perception tasks has not been fully explored. In this work, we propose DiffBEV, a novel framework that utilizes conditional DPM to improve quality of the BEV feature and push the boundary of BEV perception. In DiffBEV, the depth distribution or the BEV feature obtained from the view transformer is the input of conditional DPM. DiffBEV explores the potential of conditional diffusion model and progressively refines the noisy BEV feature. Then, the cross-attention module is proposed to fuse the fine-grained output of conditional diffusion model and the original BEV feature. This module adaptively builds the content relationship between the generated feature and the source BEV content, which helps to obtain a more precise and compact perception result.

DiffBEV is an end-to-end framework and can be easily extended by altering task-specific decoders. In this paper, we evaluate the performance of BEV semantic segmentation on standard benchmarks, \textit{i.e.} nuScenes \cite{caesar2020nuscenes}, KITTI Raw \cite{geiger2012we}, KITTI Odometry \cite{behley2019semantickitti}, and KITTI 3D Object \cite{geiger2012we}. DiffBEV achieves a \textbf{25.9\%} mIoU on the nuScenes benchmark, which is \textbf{6.2\%} higher than previous best-performing approaches. DiffBEV outperforms other methods in the segmentation of drivable area, pedestrian crossing, walkway, and car by a substantial margin (\textbf{+5.0\%}, \textbf{+10\%}, \textbf{+6.7\%}, and \textbf{+11.6\%} IoU scores). Qualitative visualization results show that DiffBEV presents more clear edges than existing approaches. Furthermore, we compare the performance of 3D object detection on the popular nuScenes benchmark with other modern 3D detectors. Without bells and whistles, DiffBEV offers benefits to 3D object detection and provides approximately 1\% NDS improvement on nuScenes. DiffBEV achieves leading performance both in BEV semantic segmentation and 3D object detection.


Our contributions can be summarized into three folds as follows.

(1) To the best of our knowledge, DiffBEV is the first work that utilizes conditional DPM to assist multiple autonomous driving perception tasks in BEV. Furthermore, DiffBEV needs no extra pre-training stage and is optimized in an end-to-end manner along with downstream tasks.

(2) The conditional DPM and the attentive fusion module are proposed to refine the original BEV feature in a progressive way, which can be seamlessly extended to different perspective view transformers, \textit{e.g.} VPN \cite{pan2020cross}, LSS \cite{philion2020lift}, PON \cite{roddick2020predicting}, and PYVA \cite{yang2021projecting}.

(3) Extensive experiments on multiple benchmarks demonstrate that DiffBEV achieves state-of-the-art performance and is effective in semantic segmentation and 3D object detection. DiffBEV achieves a \textbf{25.9\%} mIoU on the nuScenes dataset, which outperforms previous best-performing approach \cite{philion2020lift} by a substantial margin, \textit{i.e.} 6.2\% mIoU. 

\section{Related Works} \label{sec:2}

\subsection{Diffusion Model}
Diffusion models are widely used in Artificial Intelligence Generated Content (AIGC), which are of great importance in generative models. Diffusion models have illustrated their power in image generation \cite{latent-diff,tackle-gan,diff-prior,diff-var}, detection \cite{diffdet}, segmentation \cite{diffpanoptic,segdiff,ddpm-segmentation}, image-to-image translation \cite{denoising,ilvr}, super resolution \cite{diff-super}, image inpainting \cite{diff-inpaint}, image editing \cite{diff-edit}, text-to-image \cite{diff-text-1,diff-text-2,diff-text-3}, video generation \cite{diff-video-1,diff-video-2}, point cloud \cite{diff-point-1,diff-point-2,diff-point-3}, and human motion synthesis \cite{diff-motion-1,diff-motion-2}.

DDPM-Segmentation \cite{ddpm-segmentation} is the first work to apply the diffusion model to semantic segmentation, which pre-trains a diffusion model and then trains classifiers for each pixel. But the two-stage paradigm, \textit{i.e.} pre-training and fine-tuning, costs much training time, which is harmful to model efficiency. DiffusionInst \cite{diffinst} applies the diffusion model to instance segmentation. A generalist framework \cite{diffpanoptic} leverages the diffusion model to generate results of panoptic segmentation. To this end, we are motivated to further explore the potential of employing the diffusion model to generate a high-quality representation for BEV perception tasks. Compared with DDPM-Segmentation \cite{ddpm-segmentation}, DiffBEV is a generalist end-to-end framework, which doesn't need an extra pre-training stage and can be optimized along with downstream tasks.

\subsection{BEV Semantic Segmentation}

BEV semantic segmentation is a fundamental and crucial vision task in BEV scene understanding and serves as the cornerstone of path planning and controlling. VPN  \cite{pan2020cross} and PYVA \cite{yang2021projecting} present the layout of static or dynamic objects through learnable fully connected layers and attention mechanisms, respectively. LSS \cite{philion2020lift} takes advantage of camera parameters to lift image-view features to BEV and is widely applied in modern 3D detectors. HFT \cite{HFT} presents an approach to leverage the strengths of both camera parameter-free methods and camera parameter-based methods. CVT \cite{cvt} extracts the content from surrounding-view images and achieves a simple yet effective design. GitNet \cite{gitnet} follows a two-stage paradigm, improving the segmentation performance by geometry-guided pre-alignment module and ray-based transformer. However, these works suffer from defective factors, such as distortion caused by inaccurate camera parameters. In DiffBEV, we propose a conditional diffusion model to refine the distorted features and improve the performance of previous methods for BEV semantic segmentation.

\subsection{3D Object Detection}
3D object detection is a prevailing research topic for researchers in autonomous driving. FCOS3D \cite{fcos3d} proposes 3D centerness and learns the 3D attributes. PGD \cite{pgd} explores the geometric relationship of different objects and improves depth estimation. PETR \cite{petr} projects the camera parameters of multi-view images into 3D positional embeddings. BEVDet \cite{bevdet} shows the positive effects of data augmentation in image view and BEV. BEVDet4D \cite{bevdet4d} explores both the spatial and temporal content to improve the performance of 3D object detection. BEVDepth \cite{bevdepth} exploits the explicit depth supervision of multi-view images and further pushes the boundary of 3D object detection. BEVerse \cite{beverse} proposes a unified framework that jointly handles the tasks of 3D object detection, map construction, and motion prediction. In our work, we further exploit the ability of the conditional diffusion model to handle the task of 3D object detection.

\section{Approach}

\begin{figure*}
    \centering
    \includegraphics[height=8cm]{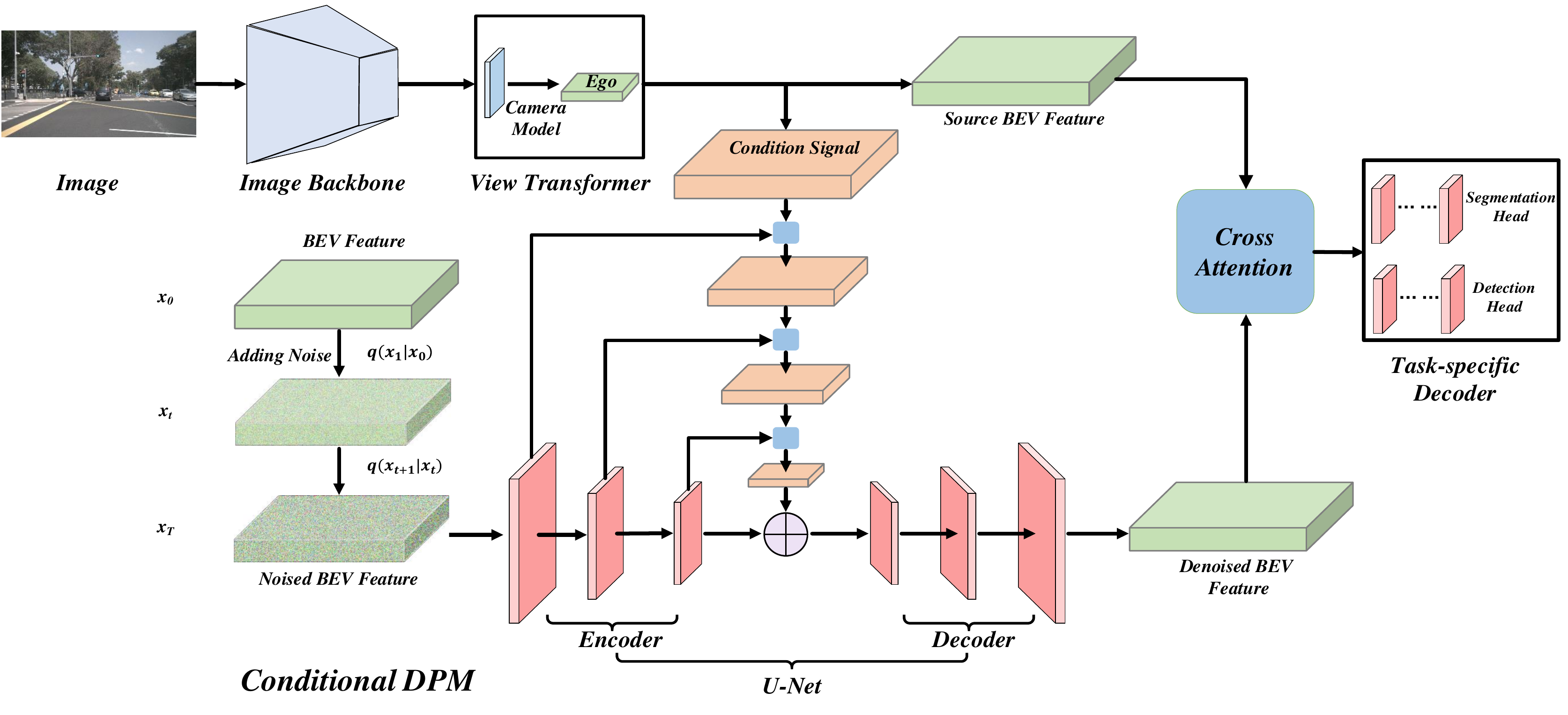}
    \caption{Overall architecture of DiffBEV. DiffBEV is comprised of the image backbone, view transformer, conditional diffusion model, cross-attention module, and task-specific decoder. By flexibly changing the task-specific decoder, DiffBEV can be easily extended to different downstream tasks, such as segmentation and 3D object detection.}
    \label{fig:architecture}
\end{figure*}
\subsection{Framework Overview}
Fig. \ref{fig:architecture} shows the overall architecture of DiffBEV, which comprises of image view backbone, view transformer, conditional diffusion model, cross-attention module, and task-specific decoder. DiffBEV doesn't require an independent stage of pre-training and is trained in an end-to-end manner.

The image view backbone extracts the image features and the view transformer lifts the image-view features to BEV. Conditional diffusion model refines noisy samples and generates high-quality semantic feature. Cross-attention module is in charge of merging BEV feature and the output of conditional diffusion model. Finally, a task-specific decoder is applied for some downstream BEV perception tasks, such as segmentation and 3D object detection. In practice, LSS \cite{philion2020lift} is adopted as the default view transformer in our implementation.

\subsection{Conditional Diffusion Probability Model}\label{sec:conditional_dpm}
\subsubsection{Diffusion Probability Model}
\ 
\quad We formulate the conditional diffusion probability model in this section. The feature generated by the view transformer is treated as the condition of diffusion model. Noise $x_T$ obeys standard normal distribution $\mathcal{N}(0, \textit{I})$. Diffusion model transforms the noise $x_T$ to the original sample $x_0$ in a progressive way. We denote the variance at step $t (0\leqslant t\leqslant T)$ as $\beta_t$. 

The forward process of the conditional diffusion model is presented as follows.
\begin{equation}
    q(x_t|x_{t-1}) \sim \mathcal{N}(x_t;\sqrt{1-\beta_t}x_{t-1},\beta_t I)
\end{equation}

For convenience, we denote a series of constant.
\begin{equation}
    \alpha_t = 1-\beta_t, \bar{\alpha_t} = \prod^t_{s=1}\alpha_s
\end{equation}

The noisy sample at step $t$ is transformed from the input data $x_0$ by Eq. \ref{forward_noise}.
\begin{equation}
    q(x_t|x_0) \sim \mathcal{N}(x_t;\sqrt{\bar{\alpha_t}}x_{t-1},(1-\alpha) I)
    \label{forward_noise}
\end{equation}
\begin{equation}
    x_t \sim \sqrt{\bar{\alpha_t}}x_0+\sqrt{1-\bar{\alpha_t}}\epsilon, \rm{where} \quad \epsilon \sim \mathcal{N}(0,\textit{I)}
\end{equation}
$\Sigma_{\theta}(x_t,t)$ is the covariance predictor and $\epsilon_{\theta}(x_t,t)$ is the denoising model. In our experiments, a typical variance of UNet \cite{medsegdiff} is used as the denoising network. In the denoising process, the diffusion model progressively refines the noisy sample $x_t$. The reverse diffusion process is written as Eq. \ref{reverse_noise}.
\begin{equation}
    p_{\theta}(x_{t-1}|x_t) \sim \mathcal{N}(x_{t-1};\mu_{\theta}(x_t,t),\Sigma_{\theta}(x_t,t))
    \label{reverse_noise}
\end{equation}

\subsubsection{The Design of Condition}
\
\indent In practice, there are three types of conditions $x_{cond}$ to choose: (1) The original BEV feature from the view transformer ($F^{O-BEV}\in\mathbb{R}^{C \times H\times W}$); (2) The semantic feature learned from the depth distribution ($F^{S-BEV}\in\mathbb{R}^{C \times H \times W}$); (3) The element-wise sum of $F^{O-BEV}$ and $F^{S-BEV}$.

The view transformer lifts the image-view feature to BEV space, obtaining the original BEV feature $F^{O-BEV}$. For each point, the view transformer estimates the distribution on different predefined depth ranges and generates the corresponding depth distribution $F^d\in \mathbb{R}^{c \times h \times w}$. We employ a $1 \times 1$ convolutional layer to convert the channel and interpolate $F^d$ into $F^{S-BEV}$, which has the same size as $F^{O-BEV}$. 

The above three conditions are features in BEV space, where we add gaussian noise. By denoising samples progressively, we hope the conditional diffusion model helps to learn the fine-granularity content of objects, such as precise boundary and highly detailed shape. We strictly follow the standard DPM model to add BEV noise, while the difference is that we employ condition-modulated denoising, which is shown in Fig. \ref{fig:architecture}. 

Given noisy BEV feature $x_t$ and condition $x_{cond}$ at time step $t$, $x_t$ is further encoded and interacts with $x_{cond}$ through element-wise multiplication. To alleviate the computational burden, we set a flexible choice for the encoding mechanism of noisy BEV feature $x_t$, \textit{i.e.} the self-attention mechanism or a simple convolutional layer, which will be discussed in Section \ref{ablations}. A UNet-style structure, whose components include an encoder and a decoder, serves as the denoising network $\epsilon_{\theta}(x_t,t)$.

\subsection{Cross-Attention Module}
\begin{figure}
    \centering
    \includegraphics[height=4.0cm]{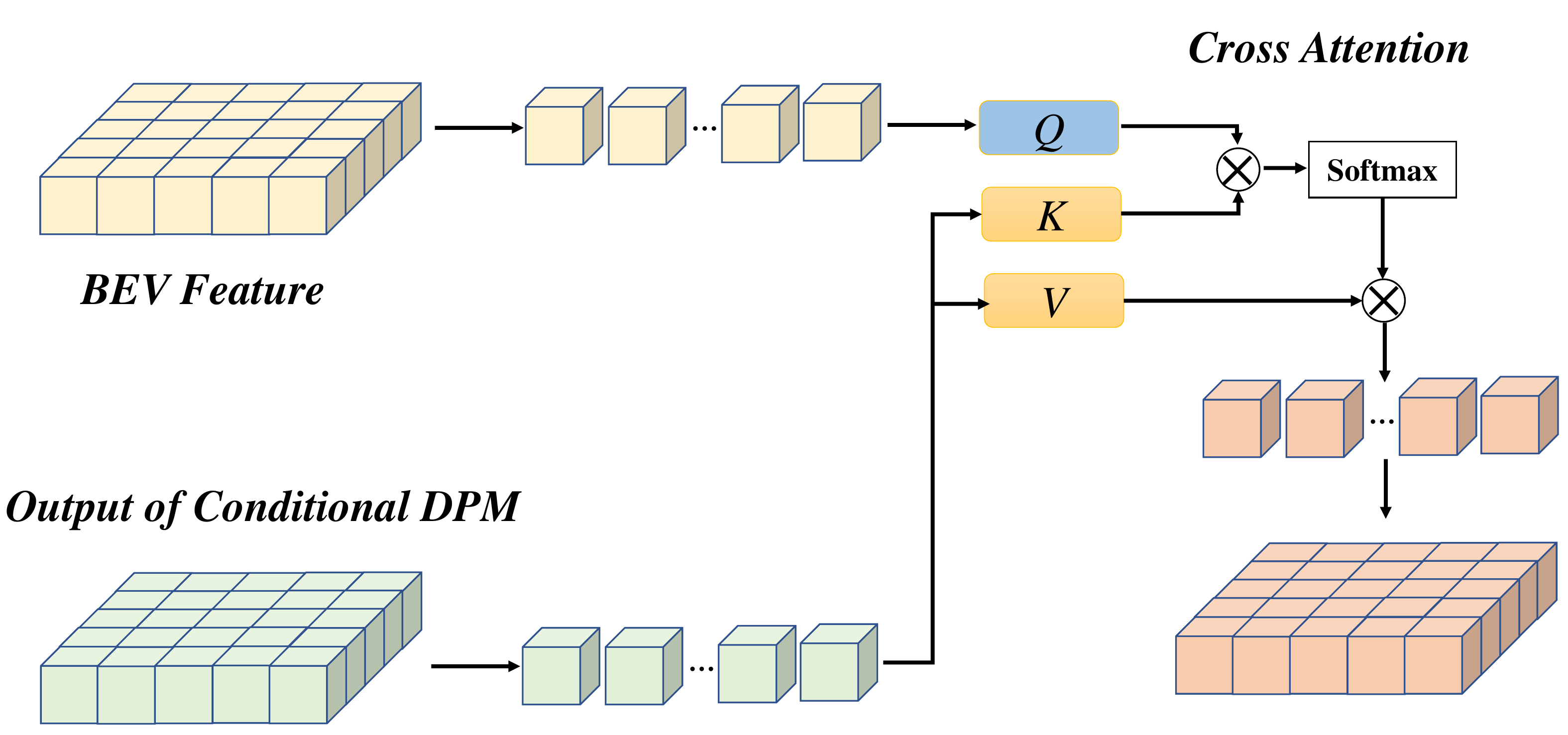}
    \caption{Overall network structure of the cross-attention module.}
    \label{fig:attentive module}
\end{figure}
After obtaining the output of conditional diffusion model, we design a cross-attention module ($CA$) to refine the original BEV feature, which is shown in Fig. \ref{fig:attentive module}. 

Specifically, the output of the conditional diffusion model is treated as the source of $K$ and $V$, while the original BEV feature from the perspective view transformer is projected into $Q$. The cross-attention process of the two-stream features is formulated as:
\begin{equation}
    \begin{aligned}
        \mathrm{CA}(Q, K, V) &= \mathrm{Attn}(QW^Q_i, KW^K_i, VW^V_i)W^{Out},\\
        \textrm{Attn}\left(Q,K,V\right) &= \textrm{softmax}\left(\frac{QK^T}{\sqrt{d_k}}\right)V.
    \end{aligned}
\end{equation}
$Q$, $K$, and $V$ are linearly mapped to calculate the attention matrix $\mathrm{Attn}$, where $W^Q_i$, $W^K_i$, $W^V_i$ are the projection layers with the shape of $\mathbb{R}^{d_{model} \times d_q}$, $ \mathbb{R}^{d_{model} \times d_k}$, $\mathbb{R}^{d_{model} \times d_v}$. Then, the refined BEV feature is obtained from the output layer $W^{Out} \in \mathbb{R}^{d_v \times d_{model}}$, which aims to facilitate the downstream tasks to learn better.

\subsection{Training Loss}

\textbf{Depth Loss.} Given the intrinsic parameter matrix $K_i \in \mathbb{R}^{3 \times 3}$, rotation matrix $R_i \in \mathbb{R}^{3 \times 3}$, and translation matrix $t_i \in \mathbb{R}^{3}$, we introduce a depth loss $\mathcal{L}_{depth}$ to assist model training. The depth loss is defined as the binary cross entropy (BCE) between the predicted depth map $D_i$ and $D_{i}^{*}$. The specific process is expressed as:
\begin{equation}
\begin{aligned}
    P_i = K_i\left(R_i P+t_i\right), D_i^* = one\_hot(P_i),\\
    \mathcal{L}_{depth} = \mathrm{BCE}(D_i^*,D_i)\\
\end{aligned}
\end{equation}

\textbf{Diffusion Loss.} We denote the gaussian noise at time step $t$ as $\bar{z}_t$. Please refer to Section \ref{sec:conditional_dpm} for the meaning of the rest symbols. The diffusion loss $\mathcal{L}_{diff}$ is defined as:
\begin{equation}
    \mathcal{L}_{diff} = \mathbb{E}[||\bar{z}_t-\Sigma_{\theta}(\sqrt{\bar{\alpha_t}}x_0+\sqrt{1-\bar{\alpha_t}}\bar{z}_t,t)||^2]
\end{equation}

\textbf{Task-specific Training Loss.} The training loss for segmentation and detection can be written as Eq. \ref{training_loss}. In practice, we empirically set the loss weights $\lambda_1 = 10$ and $\lambda_2 = 1$. We introduce the details of segmentation loss $\mathcal{L}_{wce}$ and detection loss $\mathcal{L}_{detect}$ in the supplementary material.
\begin{equation}
\begin{aligned}
    \mathcal{L}_{seg} &= \mathcal{L}_{wce}+\lambda_1 \mathcal{L}_{depth}+\lambda_2 \mathcal{L}_{diff}\\
    \mathcal{L}_{det} &= \mathcal{L}_{detect}+\lambda_1 \mathcal{L}_{depth}+\lambda_2 \mathcal{L}_{diff}\\
    \label{training_loss}
\end{aligned}
\end{equation}

\subsection{Task-specific Decoder}
As a general framework for BEV perception, DiffBEV can reason about different downstream tasks by altering the task-specific decoder. We adopt a residual-style decoding head for the semantic segmentation task, which consists of 8 convolutional blocks and a fully connected (FC) layer. Each convolutional block has a convolution layer, followed by batch normalization (BN) and a rectified linear unit (ReLU) layer. As for the 3D object detection task, the classification and regression heads are composed of several convolution layers respectively. Please refer to CenterPoint \cite{yin2021center} for more structure details of the 3D detection decoder.

\section{Experiment}
\subsection{Datasets}
We compare the performance of DiffBEV with the existing methods on four different benchmarks, \textit{i.e.} nuScenes \cite{caesar2020nuscenes}, KITTI Raw \cite{geiger2012we}, KITTI Odometry \cite{behley2019semantickitti}, and KITTI 3D Object \cite{geiger2012we}. As a popular benchmark in autonomous driving, nuScenes \cite{caesar2020nuscenes} dataset is collected by six surrounding cameras and one LiDAR, which includes multi-view images and point cloud of 1,000 scenes. KITTI Raw \cite{geiger2012we} and KITTI Odometry \cite{behley2019semantickitti} provide the images and BEV ground truth of the static road layout, while KITTI 3D Object \cite{geiger2012we} provides the images and labels for dynamic vehicles.

By flexibly leveraging different task-specific decoders, DiffBEV can be extended to various downstream tasks. In this work, extensive experiments are conducted on the BEV semantic segmentation and 3D object detection tasks. 

\subsection{Implementation Details}

We train all semantic segmentation models using the AdamW optimizer \cite{loshchilov2017fixing} with learning rate and weight decay as 2e-4 and 0.01. Two NVIDIA GeForce RTX 3090 are utilized and the mini-batch per GPU is set to 4 images. The input resolution is 800 $\times$ 600 for nuScenes and 1024 $\times$ 1024 for KITTI datasets. The total training schedule includes 20,000 iterations (200, 000 iterations for nuScenes) and the warm-up strategy \cite{goyal2017accurate} gradually increases the learning rate for the first 1,500 iterations. Then, a cyclic policy \cite{yan2018second} linearly decreases the learning rate from 2e-4 to 0 during the remainder training process. For 3D object detection, we follow the implementation details of BEVDet \cite{bevdet}.

For the image backbone, the SwinTransformer \cite{liu2021swin} is initialized with the weights pre-trained on the ImageNet \cite{Russakovsky2015ImageNetLS} dataset. The model structures of VPN \cite{pan2020cross}, PON \cite{roddick2020predicting}, LSS \cite{philion2020lift}, and PYVA \cite{yang2021projecting} are the same as the original paper. 
In addition, we mainly follow the methods of the BEVDet \cite{bevdet} family to achieve 3D object detection. The training and testing details are consistent with \cite{bevdet} and \cite{bevdet4d}. Last but not least, there is no extra pre-training stage for the conditional diffusion probability model, which can be optimized in an end-to-end manner along with the downstream tasks.

\subsection{BEV Semantic Segmentation}

\begin{table*}[t!]
\centering
\caption{Intersection over Union scores (\%) of hybrid scene layout estimation on the nuScenes \textbf{val} dataset.}
\label{nuscenes}
\begin{tabular}{@{}c|cccccccccccccc|c@{}}
\hline
Method & \rot{Drivable} & \rot{Ped. crossing} & \rot{Walkway} & \rot{Carpark} & \rot{Car} & \rot{Truck} & \rot{Bus} & \rot{Trailer} & \rot{Constr. veh.} & \rot{Pedestrian} & \rot{Motorcycle} & \rot{Bicycle} & \rot{Traf. Cone} & \rot{Barrier} & \rot{Mean} \\
\hline
IPM & 40.1 & - & 14.0 & - & 4.9 & - & 3.0 & - & - &  0.6 & 0.8 & 0.2 & - & - & - \\
Unproj. & 27.1 & - & 14.1 & - & 11.3 & - & 6.7 & - & - & 2.2 & 2.8 & 1.3 & - & - & -  \\
VED \cite{lu2019monocular}  & 54.7 & 12.0 & 20.7 & 13.5 & 8.8 & 0.2 & 0.0 & 7.4 & 0.0 & 0.0 & 0.0 & 0.0 & 0.0 & 4.0 & 8.7 \\
PYVA \cite{yang2021projecting} & 56.2 & 26.4 & 32.2 & 21.3 & 19.3 & 13.2 & 21.4 & 12.5 & 7.4 & 4.2 & 3.5 & 4.3 & 2.0 & 6.3 & 16.4 \\ 
VPN \cite{pan2020cross} & 58.0 & 27.3 & 29.4 &  12.9 & 25.5 & 17.3 & 20.0 & 16.6 & 4.9 & 7.1 & 5.6 & 4.4 & 4.6 & 10.8 & 17.5 \\ 
PON \cite{roddick2020predicting} & 60.4 & 28.0 & 31.0 &  18.4 & 24.7 & 16.8 & 20.8 & 16.6 & 12.3 & 8.2 & 7.0 & 9.4 & 5.7 & 8.1 & 19.1 \\ 
LSS \cite{philion2020lift} & 55.9 & 31.3 & 34.4 & 23.7 & 27.3 & 16.8 & 27.3 & 17.0 & 9.2 & 6.8 & 6.6 & 6.3 & 4.2 & 9.6 & 19.7 \\
\hline
DiffBEV-BEV & 65.3 & 40.2 & 41.0 & 27.2 & 37.9 & 21.3 & 32.9 & 20.5 & 7.6 & 9.2 & 13.7 & 13.1 & 7.2 & 16.0 & 25.2 \\
DiffBEV-DepBEV & 64.9 & 39.7 & 40.7 & 27.7 & 37.7 & 22.3 & 32.5 & 21.4 & \textbf{12.7} & 9.2 & 13.3 & 12.8 & 6.6 & 15.9 & 25.5 \\
DiffBEV-Dep & \textbf{65.4} & \textbf{41.3} & \textbf{41.1} & \textbf{28.4} & \textbf{38.9} & \textbf{23.1} & \textbf{33.7} & \textbf{21.1} & 8.4 & \textbf{9.6} & \textbf{14.4} & \textbf{13.2} & \textbf{7.5} & \textbf{16.7} & \textbf{25.9} \\
\hline
\end{tabular}
\end{table*}

\textbf{Evaluation on the nuScenes benchmark.} In this part, we compare the effectiveness of DiffBEV with other approaches on the pixel-wise segmentation task. Both the layout of static objects and dynamic objects are estimated on the nuScenes benchmark.

As illustrated in Tab. \ref{nuscenes}, we report the segmentation performance of DiffBEV and some advanced methods described in Section \ref{sec:2}. It can be seen that the previous state-of-the-art method LSS \cite{philion2020lift} is good at predicting static objects with wide coverage, such as the drivable area, walkway, and pedestrian crossing, compared to the car, pedestrian, bicycle, \textit{etc.} This is because dynamic objects usually occupy fewer pixels and appear less frequently in BEV. A similar performance can also be observed from PYVA \cite{yang2021projecting} and PON \cite{roddick2020predicting}, which achieve a comparable accuracy in the drivable area class but perform worse in the rare class, such as truck, bus, and trailer.

In contrast, DiffBEV has a remarkable improvement in the Intersection over Union (IoU) score of both static and dynamic objects. As listed in Tab. \ref{nuscenes}, we design three varieties according to the condition. The condition of Diff-BEV, Diff-Dep, and Diff-DepthBEV comes from the original BEV feature ($F^{O-BEV}$), conditional features learned from the depth distribution ($F^{S-BEV}$), and the element-wise sum of $F^{O-BEV}$ and $F^{S-BEV}$, respectively. DiffBEV-Dep leads the performance in most classes and achieves a \textbf{25.9\%} mIoU score, which is \textbf{6.2\%} higher than previous best-performing approach \cite{philion2020lift}. In particular, DiffBEV improves the segmentation accuracy of the drivable area, pedestrian crossing, walkway, and car by a substantial margin (\textbf{+5.0\%}, \textbf{+10.0\%}, \textbf{+6.7\%}, and \textbf{+11.6\%} IoU scores), which are crucial classes for the safety of autonomous driving systems. We attribute this improvement to that the conditional DPM reduces noises and complements more spatial information about objects of interest. DiffBEV significantly improves the pixel-wise perception accuracy of the model in both high-frequency classes and sparsely distributed classes. Please refer to Section \ref{sec:vis} for a more intuitive analysis and explanation.
 
\textbf{Evaluation on KITTI Raw, KITTI Odometry, and KITTI 3D Object benchmark.} Tab.~\ref{tab:Kitti raw and odometry} reports the quantitative results of static scene layout estimation on KITTI Raw and KITTI Odometry datasets. The performance comparison on KITTI 3D Object dataset shows the segmentation results for dynamic vehicles. Three varieties of DiffBEV obtain higher mIoU and mAP scores than existing methods. For example, DiffBEV-Dep surpasses the second-best model PYVA \cite{yang2021projecting} by 0.71\%, 1.51\%, and 7.97\% mIoU on KITTI Raw, KITTI Odometry, and KITTI 3D Object dataset, which achieves state-of-the-art perception accuracy consistently on all evaluation benchmarks.
\begin{table}[t!]
\centering
\caption{Segmentation performance of static scene layout estimation on KITTI Raw and KITTI Odometry, and dynamic scene layout estimation on KITTI 3D Object. mIoU (\%) and mAP (\%) metrics are reported.} 
\label{tab:Kitti raw and odometry}
\resizebox{\linewidth}{!}{
\begin{tabular}{@{}c|cc|cc|cc@{}}
\hline
KITTI & \multicolumn{2}{|c|}{Raw} & \multicolumn{2}{|c|}{Odometry}& \multicolumn{2}{|c}{3D Object}\\
\hline
Method & mIoU & mAP & mIoU & mAP & mIoU & mAP\\
\hline
OFT \cite{3D3}&-&-&-&-&25.34&34.69\\
MonoOcc \cite{lu2019monocular} & 58.41 & 66.01 & 65.74 & 67.84&20.45&22.29 \\
Mono3D \cite{chen2016monocular} & 59.58 & 79.07 & 66.81 & 81.79 &17.11&26.62\\
VPN \cite{pan2020cross}& 64.65 & 78.20 & 78.16 & 84.73 &26.52&35.54\\
PYVA \cite{yang2021projecting} & 65.70 & 81.62 & 78.19 & 85.55&29.11&36.86  \\
PON \cite{roddick2020predicting} & 60.47 &  77.45 &  70.92 &  76.27&26.78&44.50 \\
\hline
DiffBEV-BEV & 66.19 & 81.08 & 79.48 & 88.30 & 36.76 & 52.81 \\
DiffBEV-DepBEV & 66.40 & 81.89 & 79.58 & 88.44 & \textbf{37.08} & \textbf{53.96} \\
DiffBEV-Dep & \textbf{66.41} & \textbf{81.91} & \textbf{79.70} & \textbf{89.68} & 36.99 & 53.61 \\  
\hline
\end{tabular}}
\end{table}

\subsection{3D Object Detection}
\begin{table*}[t]
  \centering
  \caption{3D object detection performance of different paradigms on the nuScenes \texttt{val} set. Tiny means tiny Swin Transformer.}
  	\resizebox{0.84\linewidth}{!}{
    \begin{tabular}{l|crrc|cccccc|c}
    \hline
    Methods   &Image Size    &\textbf{mAP}$\uparrow$ & mATE$\downarrow$  & mASE$\downarrow$   & mAOE$\downarrow$  & mAVE$\downarrow$  &  mAAE$\downarrow$ & \textbf{NDS}$\uparrow$  \\
    \hline
    CenterNet\cite{centernet}  &-        & 0.306         & 0.716             & 0.264              & 0.609             & 1.426             & 0.658             & 0.328           \\
    FCOS3D \cite{fcos3d}       &1600$\times$900    & 0.295         & 0.806             & 0.268              & 0.511             & 1.315             & \textbf{0.170}    & 0.372           \\
    DETR3D \cite{detr3d}       &1600$\times$900    & 0.303         & 0.860             & 0.278              & 0.437             & 0.967             & 0.235             & 0.374           \\
    PGD \cite{pgd}             &1600$\times$900    & 0.335         & 0.732             & 0.263              & \textbf{0.423}             & 1.285             & 0.172             & 0.409           \\
    PETR-R50  \cite{petr}      &1056$\times$384    & 0.313         & 0.768             & 0.278              & 0.564             & 0.923             & 0.225             & 0.381           \\
    PETR-R101  \cite{petr}      &1408$\times$512    & 0.357         & 0.710             & 0.270              & 0.490             & 0.885             & 0.224             & 0.421           \\
    PETR-Tiny \cite{petr}      &1408$\times$512    & \textbf{0.361}         & 0.732             & 0.273              & 0.497             & 0.808             & 0.185             & 0.431           \\
    \hline
    BEVDet-Tiny\cite{bevdet}   &704$\times$256     & 0.310         & 0.681             & 0.273              & 0.570             & 0.933             & 0.223             & 0.387           \\
    BEVDet-Tiny \cite{bevdet}+DiffBEV  &704$\times$256  & 0.315       & 0.660  & 0.265 & 0.567  & 0.878  & 0.219 & 0.398\\
    \hline
    BEVDet4D-Tiny \cite{bevdet4d}      &704$\times$256 & 0.338         & 0.672     & 0.274  & 0.460             & 0.337             & 0.185             & 0.476           \\
    BEVDet4D-Tiny \cite{bevdet4d}+DiffBEV  &704$\times$256 & 0.344 & \textbf{0.652}  &  \textbf{0.262}  & 0.453 &  \textbf{0.312} & 0.176 & \textbf{0.486}\\
    \hline
    \end{tabular}
    }
  \label{tab:nuscenes-det}
\end{table*}
We conduct 3D object detection experiments on the nuScenes benchmark and Tab.~\ref{tab:nuscenes-det} reports the official evaluation metrics: mean Average Precision (mAP), Average Translation Error (ATE), Average Scale Error (ASE), Average Orientation Error (AOE), Average Velocity Error (AVE), Average Attribute Error (AAE), and NuScenes Detection Score (NDS). Note that we select LSS \cite{philion2020lift} as the default view transformer, and use the semantic feature learned from the depth distribution ($F^{S-BEV}$) as the condition of DiffBEV. The data augmentations in image view and BEV are strictly consistent with that of the BEVDet \cite{bevdet} and BEVDet4D \cite{bevdet4d}.

After applying the conditional diffusion model, it can be observed that all evaluation metrics for 3D object detection are improved. This is because DiffBEV progressively refines the original BEV feature and interactively exchanges the semantic context through the cross-attention mechanism. Without bells and whistles, BEVDet \cite{bevdet} with DiffBEV raises the NDS score from 38.7\% to 39.8\%, while BEVDet4D \cite{bevdet4d} with DiffBEV raises the NDS score from 47.6\% to 48.6\%.

\subsection{Ablation Study}\label{ablations}
\subsubsection{Condition Design}
\ 
\indent In order to exploit the advantages of the conditional diffusion model, we conduct ablation experiments for different DPM conditions on the KITTI Raw dataset to estimate the layout of static roads. Specifically, there are three DPM conditions to choose, \textit{i.e.} the original BEV feature ($F^{O-BEV}$), the semantic feature learned from the depth distribution ($F^{S-BEV}$), and the element-wise sum of $F^{O-BEV}$ and $F^{S-BEV}$ ($F^{S-BEV}$ \& $F^{O-BEV}$). 

As shown in Tab. \ref{tab:ablation on condition}, no matter which condition is used, three conditions can guide the DPM to learn discriminative BEV feature. $F^{S-BEV}$ and $F^{S-BEV}$ \& $F^{O-BEV}$ achieve better modulation effects than the $F^{O-BEV}$, while the best segmentation result comes from $F^{S-BEV}$. This observation demonstrates the effectiveness of semantic feature learned from the depth distribution.

\begin{table}[t!]
\centering
\caption{Ablation study on condition design and feature fusion mechanism. The mIoU (\%) score of the basic LSS \cite{philion2020lift} on the KITTI Raw dataset is 63.38\%.}
\label{tab:ablation on condition}
\resizebox{1.0\columnwidth}{!}{\begin{tabular}{@{}c|ccc@{}}
\hline
Interaction Mechanism & $F^{S-BEV}$ & $F^{O-BEV}$ & $F^{S-BEV}$ \& $F^{O-BEV}$\\
\hline
Concat & 65.03 & 64.81 & 64.95\\
Add & 64.85 & 64.11 & 64.50 \\
Cross-Attention & \textbf{65.86} & 64.33 & 65.16 \\
\hline
\end{tabular}}
\end{table}

\subsubsection{Feature Interaction Mechanism}
\ 
\indent Another ablation study is to explore the most effective way for feature interaction. Specifically, three mechanisms, namely, concatenation (Concat), summation (Add), and cross-attention are explored in our experiments. 

As shown in each row of Tab. \ref{tab:ablation on condition}, regardless of which feature interaction mechanism is employed, DiffBEV achieves better segmentation results than the baseline model with 63.38\% mIoU. It can be seen that cross-attention can learn better BEV feature than the other two simple feature interactions, which is beneficial for the downstream perception tasks. In summary, the combination of $F^{S-BEV}$ and the cross-attention feature interaction mechanism achieves the best segmentation results, which improves 2.48\% mIoU based on LSS \cite{philion2020lift} model. If not specified, the DiffBEV model corresponds to the setting of $F^{S-BEV}$ with the cross-attention mechanism.

\newcommand{\linesep}{\vfill\vspace{0.1ex}\vfill}
\begin{figure*}[htbp]
\centering
\begin{minipage}[c]{0.202\linewidth}
\centering
\includegraphics[height=0.53\linewidth]{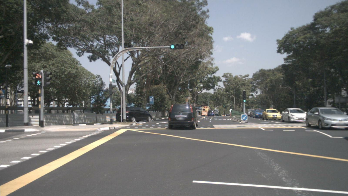}\linesep
\includegraphics[height=0.53\linewidth]{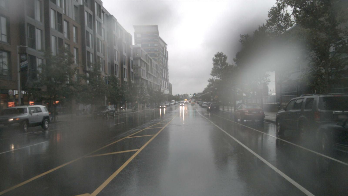}\linesep
\includegraphics[height=0.53\linewidth]{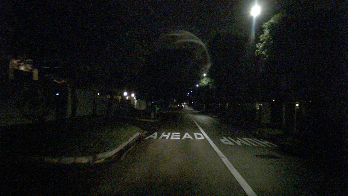}\linesep
\includegraphics[height=0.53\linewidth]{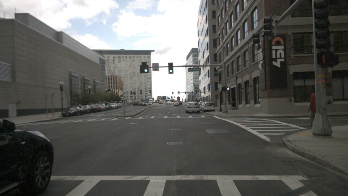}\linesep
\scriptsize{Image}\linesep
\end{minipage}
\hspace{-1.2ex}
\begin{minipage}[c]{0.121\linewidth}
\centering
\includegraphics[width=0.9\linewidth]{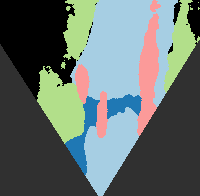}\linesep
\includegraphics[width=0.9\linewidth]{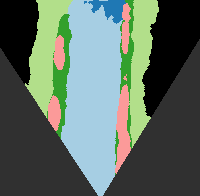}\linesep
\includegraphics[width=0.9\linewidth]{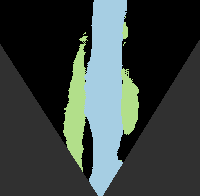}\linesep
\includegraphics[width=0.9\linewidth]{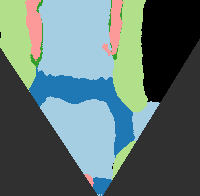}\linesep
\scriptsize{VPN\cite{pan2020cross}}\linesep
\end{minipage}
\hspace{-1.2ex}
\begin{minipage}[c]{0.121\linewidth}
\centering
\includegraphics[width=0.9\linewidth]{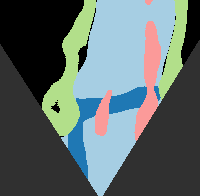}\linesep
\includegraphics[width=0.9\linewidth]{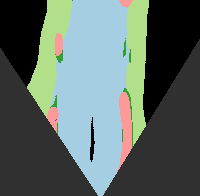}\linesep
\includegraphics[width=0.9\linewidth]{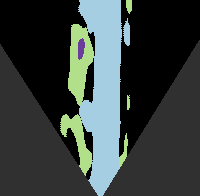}\linesep
\includegraphics[width=0.9\linewidth]{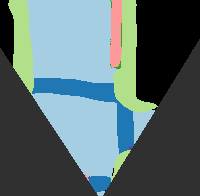}\linesep
\scriptsize{PYVA\cite{yang2021projecting}}\linesep
\end{minipage}
\hspace{-1.2ex}
\begin{minipage}[c]{0.121\linewidth}
\centering
\includegraphics[width=0.9\linewidth]{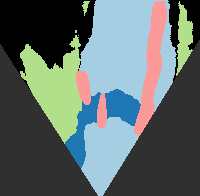}\linesep
\includegraphics[width=0.9\linewidth]{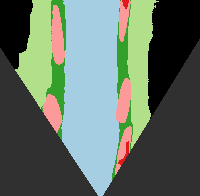}\linesep
\includegraphics[width=0.9\linewidth]{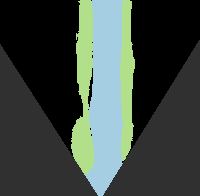}\linesep
\includegraphics[width=0.9\linewidth]{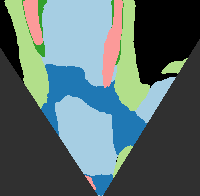}\linesep
\scriptsize{PON\cite{roddick2020predicting}}\linesep
\end{minipage}
\hspace{-1.2ex}
\begin{minipage}[c]{0.121\linewidth}
\centering
\includegraphics[width=0.9\linewidth]{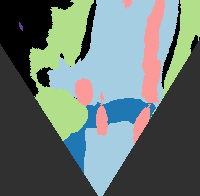}\linesep
\includegraphics[width=0.9\linewidth]{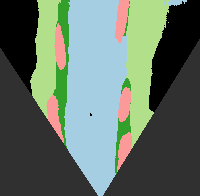}\linesep
\includegraphics[width=0.9\linewidth]{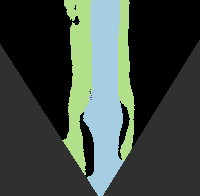}\linesep
\includegraphics[width=0.9\linewidth]{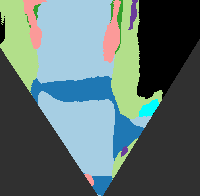}\linesep
\scriptsize{LSS\cite{philion2020lift}}\linesep
\end{minipage}
\hspace{-1.2ex}
\begin{minipage}[c]{0.121\linewidth}
\centering
\includegraphics[width=0.9\linewidth]{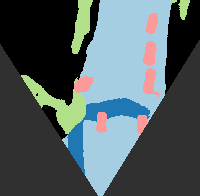}\linesep
\includegraphics[width=0.9\linewidth]{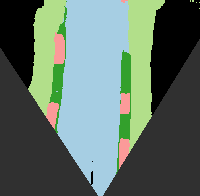}\linesep
\includegraphics[width=0.9\linewidth]{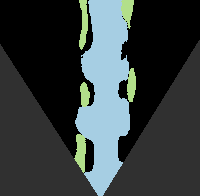}\linesep
\includegraphics[width=0.9\linewidth]{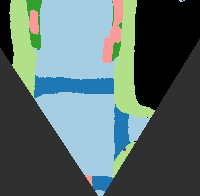}\linesep
\scriptsize{DiffBEV}\linesep
\end{minipage}
\hspace{-1.2ex}
\begin{minipage}[c]{0.121\linewidth}
\centering
\includegraphics[width=0.9\linewidth]{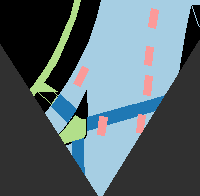}\linesep
\includegraphics[width=0.9\linewidth]{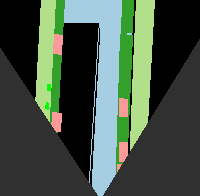}\linesep
\includegraphics[width=0.9\linewidth]{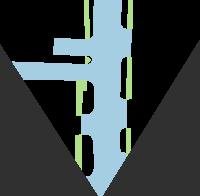}\linesep
\includegraphics[width=0.9\linewidth]{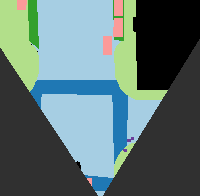}\linesep
\scriptsize{Ground Truth}\linesep
\end{minipage}
\hspace{-1.2ex}
    \caption{Qualitative segmentation results on the nuScenes benchmark. We visualize the class with the largest index $c$ which has occupancy probability $p_i > 0.5$. Black regions (outside field of view or no LiDAR returns) are ignored during evaluation. Please refer to the top legend in Fig. \ref{fig:intro} for the meaning of different colors.}
    \label{fig:vis_results}
    \vspace{-5pt}
\end{figure*}

\begin{figure*}[htbp]
\centering
\begin{minipage}[c]{0.202\linewidth}
\centering
\includegraphics[height=0.53\linewidth]{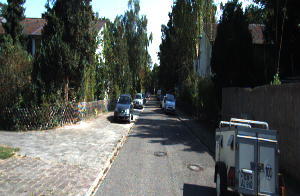}\linesep
\includegraphics[height=0.53\linewidth]{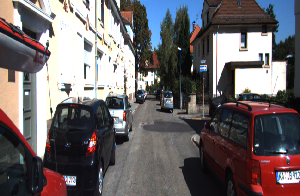}\linesep
\includegraphics[height=0.53\linewidth]{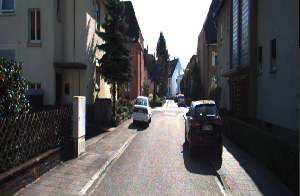}\linesep
\includegraphics[height=0.53\linewidth]{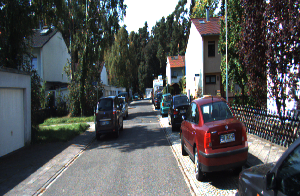}\linesep
\scriptsize{Image}\linesep
\end{minipage}
\hspace{-1.2ex}
\begin{minipage}[c]{0.121\linewidth}
\centering
\includegraphics[width=0.9\linewidth]{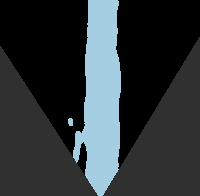}\linesep
\includegraphics[width=0.9\linewidth]{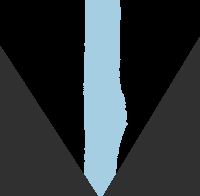}\linesep
\includegraphics[width=0.9\linewidth]{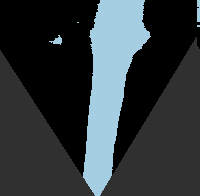}\linesep
\includegraphics[width=0.9\linewidth]{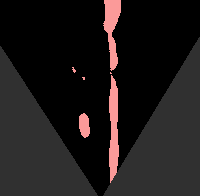}\linesep
\scriptsize{VPN\cite{pan2020cross}}\linesep
\end{minipage}
\hspace{-1.2ex}
\begin{minipage}[c]{0.121\linewidth}
\centering
\includegraphics[width=0.9\linewidth]{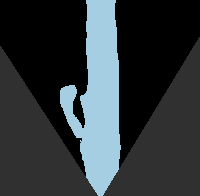}\linesep
\includegraphics[width=0.9\linewidth]{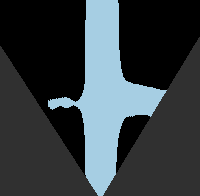}\linesep
\includegraphics[width=0.9\linewidth]{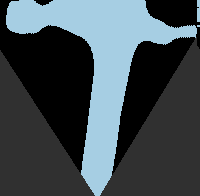}\linesep
\includegraphics[width=0.9\linewidth]{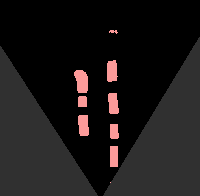}\linesep
\scriptsize{PYVA\cite{yang2021projecting}}\linesep
\end{minipage}
\hspace{-1.2ex}
\begin{minipage}[c]{0.121\linewidth}
\centering
\includegraphics[width=0.9\linewidth]{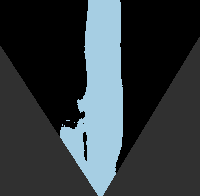}\linesep
\includegraphics[width=0.9\linewidth]{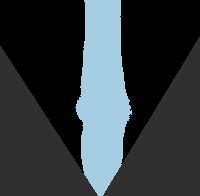}\linesep
\includegraphics[width=0.9\linewidth]{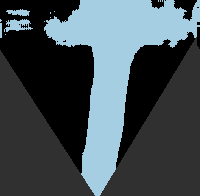}\linesep
\includegraphics[width=0.9\linewidth]{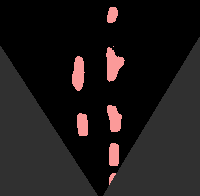}\linesep
\scriptsize{PON\cite{roddick2020predicting}}\linesep
\end{minipage}
\hspace{-1.2ex}
\begin{minipage}[c]{0.121\linewidth}
\centering
\includegraphics[width=0.9\linewidth]{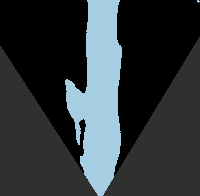}\linesep
\includegraphics[width=0.9\linewidth]{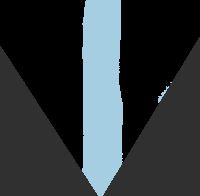}\linesep
\includegraphics[width=0.9\linewidth]{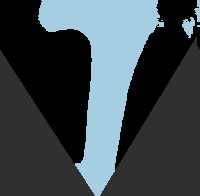}\linesep
\includegraphics[width=0.9\linewidth]{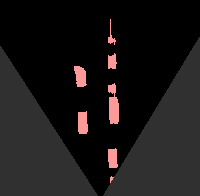}\linesep
\scriptsize{LSS\cite{philion2020lift}}\linesep
\end{minipage}
\hspace{-1.2ex}
\begin{minipage}[c]{0.121\linewidth}
\centering
\includegraphics[width=0.9\linewidth]{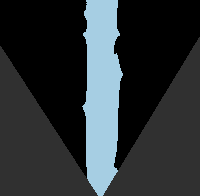}\linesep
\includegraphics[width=0.9\linewidth]{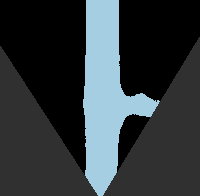}\linesep
\includegraphics[width=0.9\linewidth]{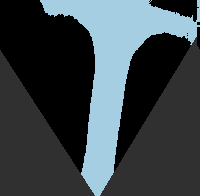}\linesep
\includegraphics[width=0.9\linewidth]{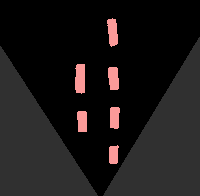}\linesep
\scriptsize{DiffBEV}\linesep
\end{minipage}
\hspace{-1.2ex}
\begin{minipage}[c]{0.121\linewidth}
\centering
\includegraphics[width=0.9\linewidth]{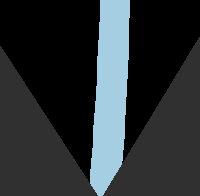}\linesep
\includegraphics[width=0.9\linewidth]{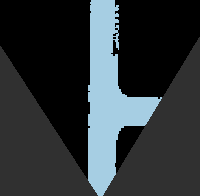}\linesep
\includegraphics[width=0.9\linewidth]{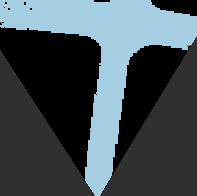}\linesep
\includegraphics[width=0.9\linewidth]{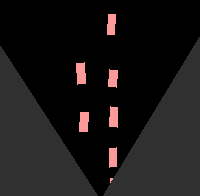}\linesep
\scriptsize{Ground Truth}\linesep
\end{minipage}
\hspace{-1.2ex}
    \caption{Qualitative results on KITTI Raw (the 1st row), KITTI Odometry (the 2nd and 3rd row), and KITTI 3D Object (the last row).}
    \label{fig:vis_results_kitti}
    \vspace{-6pt}
\end{figure*}

\subsubsection{Encoding Mechanism for Noisy BEV Samples}
\ 
\indent In this part, we explore the inner encoding mechanism in the conditional diffusion model. For the noisy BEV sample $x_t$, we calculate the self-attention semantic map or obtain the refined affinity map through a simple convolutional layer. Tab.~\ref{tab:interaction_diffusion} shows the comparison between the computational burden and segmentation performance. The DiffBEV model using self-attention mechanism achieves a higher 65.86\% mIoU and an 80.62\% mAP. By simplifying self-attention to a simple convolutional layer, the DiffBEV model achieves a 64.23\% mIoU and a 78.34\% mAP while decreases the GFLOPs from 446.81 to 433.72. 
\begin{table}[t!]
\centering
\caption{Ablation study on encoding mechanism in conditional diffusion model. The mIoU and mAP (\%) of the basic LSS \cite{philion2020lift} on the KITTI Raw dataset are 63.38\% and 77.52\%, respectively.}
\label{tab:interaction_diffusion}
\resizebox{1.0\columnwidth}{!}{\begin{tabular}{@{}c|cccc@{}}
\hline
Encoding Mechanism & \#param. & GFLOPs & mIoU & mAP\\
\hline
Conv & 78.16M & 433.72 & 64.23 & 78.34\\
Self-Attention & 78.80M & 446.81 & 65.86 & 80.62 \\
\hline
\end{tabular}}
\end{table}

\subsection{More View Transformers with DiffBEV}
In the main experiments, we adopt LSS \cite{philion2020lift} as the view transformer, which lifts the 2D image feature into BEV space. To investigate the generality of DiffBEV, we also conduct experiments on more view transformers.

As shown in Tab.~\ref{tab:more view transformer}, the model equipped with DiffBEV outperforms the version without DPM on both mIoU and mAP metrics by a significant margin. Benefited from DiffBEV, the models of VPN \cite{pan2020cross}, PYVA \cite{yang2021projecting}, and PON \cite{roddick2020predicting} raise their performances on mIoU scores (\textbf{+1.19\%}, \textbf{+1.61\%}, \textbf{+0.59\%}, respectively) and mAP scores (\textbf{+10.14\%}, \textbf{+7.01\%}, \textbf{+10.11\%}, respectively). This observation illustrates that DiffBEV is not only effective for a specific view transformer. The experimental results illustrate that DiffBEV provides a simple yet effective way to improve the quality of BEV semantic segmentation.

\begin{table}[t!]
\centering
\caption{Extension experiments of more view transformers with DiffBEV on the KITTI 3D Object dataset. The metric (\%) in the middle and right columns represent the performance without and with DiffBEV, respectively. Note that we conduct these experiments with 1024 $\times$ 1024 resolution.}
\label{tab:more view transformer}
\resizebox{0.99\columnwidth}{!}{
\begin{tabular}{@{}c|cc|cc@{}}
\hline
Model & \multicolumn{2}{c}{mIoU} & \multicolumn{2}{|c}{mAP} \\
\hline
DiffBEV & \XSolidBrush & \checkmark & \XSolidBrush  & \checkmark \\
\hline
VPN \cite{pan2020cross} & 27.02 & 28.21 (+1.19) & 35.63 & 45.77 (\textbf{+10.14})\\
PYVA \cite{yang2021projecting} & 29.22 & 30.83 (\textbf{+1.61}) & 36.97 & 43.98 (+7.01) \\
PON \cite{roddick2020predicting} & 36.49 & 37.08 (+ 0.59) & 45.51 & 55.62 (+10.11)\\
\hline
\end{tabular}}
\end{table}

\subsection{Visualization Analysis} \label{sec:vis}
In this part, we visualize some images and the corresponding semantic maps generated by different approaches, \textit{i.e.} VPN \cite{pan2020cross}, PYVA \cite{yang2021projecting}, PON \cite{roddick2020predicting}, LSS \cite{philion2020lift}, and DiffBEV.
As indicated in Fig. \ref{fig:vis_results}, previous state-of-the-art methods tend to output relatively rough predictions, \textit{i.e.} the boundaries of different objects are blurred and the categories are confused. For instance, cars that should be independent individuals are connected into a strip region and the drivable area is misclassified as the background. The predictions of the walkway and pedestrian crossing are too distorted to provide accurate location information.

Despite the complex and challenging street layouts on the nuScenes dataset, DiffBEV produces more accurate semantic maps and is able to resolve fine-grained details such as the spatial separation between neighboring vehicles, especially in the crowded autonomous driving scenarios. As shown in Fig. \ref{fig:vis_results_kitti}, similar performance advantages can also be observed on the KITTI dataset. For static roads on the KITTI Raw and KITTI Odometry datasets, DiffBEV outperforms the previous methods and provides semantic maps with more clear edges, which confirms that our method is able to effectively learn the structural information of objects.

\section{Conclusion}
In this work, we propose a novel framework, namely DiffBEV, which first applies the conditional diffusion model to BEV perception tasks. DiffBEV utilizes the BEV feature, the semantic feature learned from the depth distribution, or the sum of these two features as the condition of the diffusion model, which progressively refines the noisy samples to generate highly detailed semantic information. Then, a cross-attention module is proposed to attentively learn the interactive relationship between the output of conditional DPM and BEV feature. Extensive experiments on multiple benchmarks illustrate that DiffBEV achieves favorable performance in both semantic segmentation and 3D object detection tasks. For instance, DiffBEV obtains a 25.9\% mIoU on the nuScenes benchmark, outperforming the previous state-of-the-art method by a substantial margin. The extension studies on different view transformers also confirm the generality of DiffBEV. Considering the rapid research progress of diffusion models, we hope to further explore the potential of DiffBEV and broaden its application ranges to more BEV perception tasks.

{\small
\bibliographystyle{ieee_fullname}
\bibliography{egbib}
}

\appendix
\twocolumn[{
\centering
\section*{\Large \centering Supplementary Material of \\ 
DiffBEV: Conditional Diffusion Model for Bird’s Eye View Perception}
 \vspace{30pt}
 }]

\section{Training Loss}
\subsection{Segmentation Loss}
For the semantic segmentation task of $M$ categories, the training loss $\mathcal{L}_{wce}$ is decomposed to $M$ weighted binary classification loss.
\begin{equation} \label{eq:wce}
    \mathcal{L}_{wce} = \sum_{c=1}^{M}\frac{w_c}{N_{pos}}[-\sum_i^{N_{pos}} y_i\log p_{ci}-\sum_i^{N_{neg}}(1-y_i)\log(1-p_{ci})]
\end{equation}
where $p_{ci}$ is the predicted classification confidence of each pixel sample and the the class-wise weight $w_c$ is calculated according to the class distribution. $y_i$ stands for the semantic label of each pixel. $N_{pos}$ and $N_{neg}$ are the number of positive samples and negative samples, respectively.

\subsection{Detection Loss}
For 3D object detection task, the training loss includes classification term $\mathcal{L}_{cls}$ and regression term $\mathcal{L}_{reg}$, which can be written as:
\begin{equation}
\mathcal{L}_{detect} = \mathcal{L}_{cls}(H^*,H_{pred})+\mathcal{L}_{reg}(box^*,box)
\end{equation}
where $H_{pred}$ is the predicted heatmap, while $box$ is predicted coordinates of bounding box. $H^*$ and $box^*$ are the corresponding ground truths. The implementation details are consistent with that of CenterPoint \cite{yin2021center}.

\subsection{Depth Loss} 
We introduce the depth loss $\mathcal{L}_{depth}$ during the training stage. Given the intrinsic parameter matrix $K_i \in \mathbb{R}^{3 \times 3}$, rotation matrix $R_i \in \mathbb{R}^{3 \times 3}$, and translation matrix $t_i \in \mathbb{R}^{3}$, the depth loss $\mathcal{L}_{depth}$ is introduced to assist the model to learn depth distribution. The depth loss is defined as the binary cross entropy (BCE) between the predicted depth map $D_i$ and $D_{i}^{*}$. The ground truth of the depth map is obtained by projecting the point cloud ($P$) into $i$-th image view. The projecting process is expressed as:
\begin{equation}
\begin{aligned}
    P_i = K_i\left(R_i P+t_i\right), D_i^* = one\_hot(P_i) \\
    \mathcal{L}_{depth} = \mathrm{BCE}(D_i^*,D_i)\\
\end{aligned}
\end{equation}

\subsection{Diffusion Loss} 
Given the time step $t$, we denote the gaussian noise as $\bar{z}_t$. Please refer to the main paper for the meaning of the rest symbols. The diffusion loss $\mathcal{L}_{diff}$ is defined as:
\begin{equation}
    \mathcal{L}_{diff} = \mathbb{E}[||\bar{z}_t-\Sigma_{\theta}(\sqrt{\bar{\alpha_t}}x_0+\sqrt{1-\bar{\alpha_t}}\bar{z}_t,t)||^2]
\end{equation}

\subsection{Task-specific Training Loss} 
For specific downstream tasks, \textit{i.e.} segmentation and detection, the training loss can be written as Eq. \ref{training_loss_appendix}. In our experiments, we empirically set the loss weights $\lambda_1 = 10$ and $\lambda_2 = 1$.
\begin{equation}
\begin{aligned}
    \mathcal{L}_{seg} &= \mathcal{L}_{wce}+\lambda_1 \mathcal{L}_{depth}+\lambda_2 \mathcal{L}_{diff}\\
    \mathcal{L}_{det} &= \mathcal{L}_{detect}+\lambda_1 \mathcal{L}_{depth}+\lambda_2 \mathcal{L}_{diff}\\
    \label{training_loss_appendix}
\end{aligned}
\end{equation}

\section{More Visualization Analysis}
In this section, we visualize more images and the corresponding semantic maps generated by different methods, \textit{i.e.} VPN \cite{pan2020cross}, PYVA \cite{yang2021projecting}, PON \cite{roddick2020predicting}, LSS \cite{philion2020lift}, and DiffBEV. 

\subsection{More Qualitative Analysis on the nuScenes Dataset}
\begin{figure*}[htbp]
\centering
\begin{minipage}[c]{0.202\linewidth}
\centering
\includegraphics[height=0.53\linewidth]{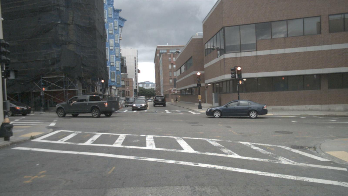}\linesep
\includegraphics[height=0.53\linewidth]{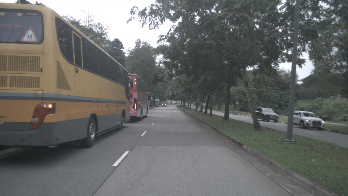}\linesep
\includegraphics[height=0.53\linewidth]{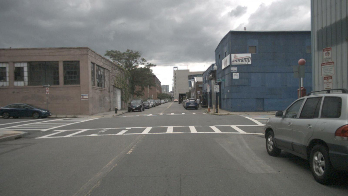}\linesep
\includegraphics[height=0.53\linewidth]{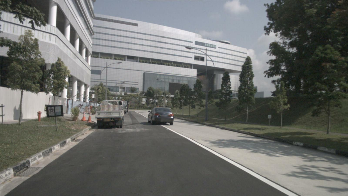}\linesep
\includegraphics[height=0.53\linewidth]{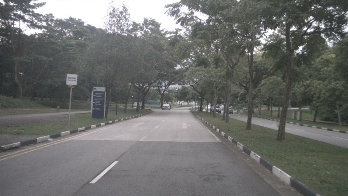}\linesep
\scriptsize{Image}\linesep
\end{minipage}
\hspace{-1.2ex}
\begin{minipage}[c]{0.121\linewidth}
\centering
\includegraphics[width=0.9\linewidth]{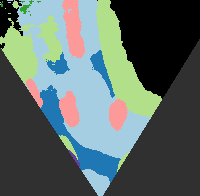}\linesep
\includegraphics[width=0.9\linewidth]{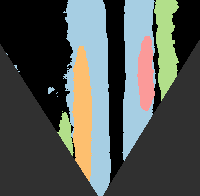}\linesep
\includegraphics[width=0.9\linewidth]{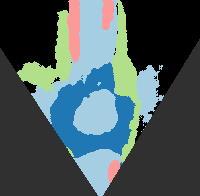}\linesep
\includegraphics[width=0.9\linewidth]{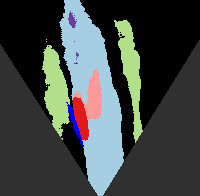}\linesep
\includegraphics[width=0.9\linewidth]{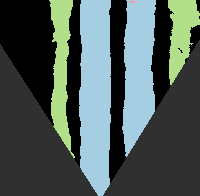}\linesep
\scriptsize{VPN\cite{pan2020cross}}\linesep
\end{minipage}
\hspace{-1.2ex}
\begin{minipage}[c]{0.121\linewidth}
\centering
\includegraphics[width=0.9\linewidth]{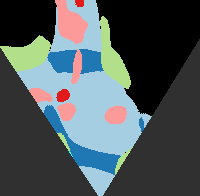}\linesep
\includegraphics[width=0.9\linewidth]{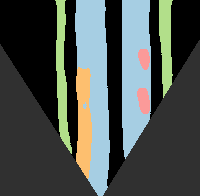}\linesep
\includegraphics[width=0.9\linewidth]{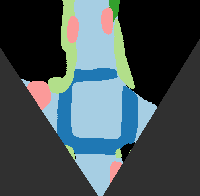}\linesep
\includegraphics[width=0.9\linewidth]{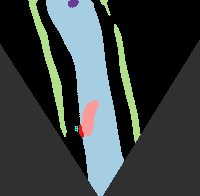}\linesep
\includegraphics[width=0.9\linewidth]{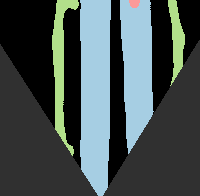}\linesep
\scriptsize{PYVA\cite{yang2021projecting}}\linesep
\end{minipage}
\hspace{-1.2ex}
\begin{minipage}[c]{0.121\linewidth}
\centering
\includegraphics[width=0.9\linewidth]{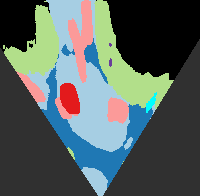}\linesep
\includegraphics[width=0.9\linewidth]{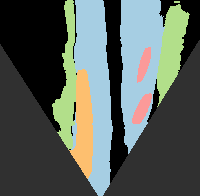}\linesep
\includegraphics[width=0.9\linewidth]{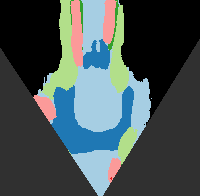}\linesep
\includegraphics[width=0.9\linewidth]{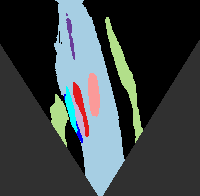}\linesep
\includegraphics[width=0.9\linewidth]{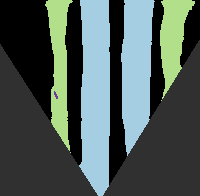}\linesep
\scriptsize{PON\cite{roddick2020predicting}}\linesep
\end{minipage}
\hspace{-1.2ex}
\begin{minipage}[c]{0.121\linewidth}
\centering
\includegraphics[width=0.9\linewidth]{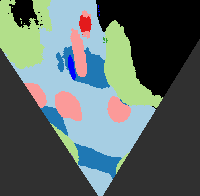}\linesep
\includegraphics[width=0.9\linewidth]{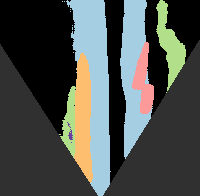}\linesep
\includegraphics[width=0.9\linewidth]{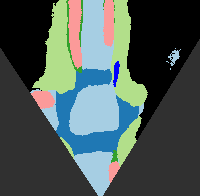}\linesep
\includegraphics[width=0.9\linewidth]{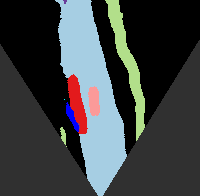}\linesep
\includegraphics[width=0.9\linewidth]{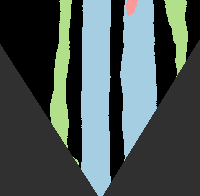}\linesep
\scriptsize{LSS\cite{philion2020lift}}\linesep
\end{minipage}
\hspace{-1.2ex}
\begin{minipage}[c]{0.121\linewidth}
\centering
\includegraphics[width=0.9\linewidth]{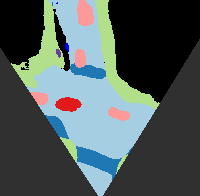}\linesep
\includegraphics[width=0.9\linewidth]{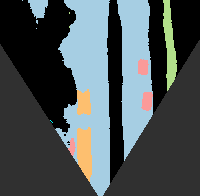}\linesep
\includegraphics[width=0.9\linewidth]{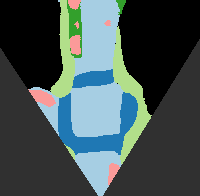}\linesep
\includegraphics[width=0.9\linewidth]{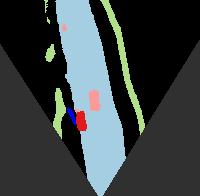}\linesep
\includegraphics[width=0.9\linewidth]{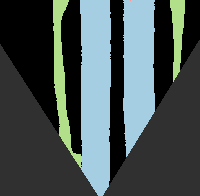}\linesep
\scriptsize{DiffBEV}\linesep
\end{minipage}
\hspace{-1.2ex}
\begin{minipage}[c]{0.121\linewidth}
\centering
\includegraphics[width=0.9\linewidth]{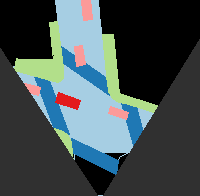}\linesep
\includegraphics[width=0.9\linewidth]{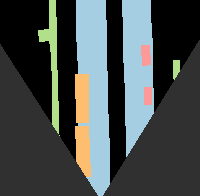}\linesep
\includegraphics[width=0.9\linewidth]{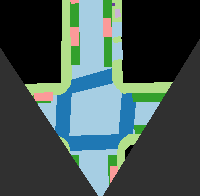}\linesep
\includegraphics[width=0.9\linewidth]{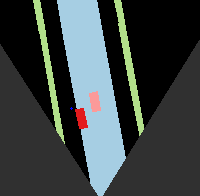}\linesep
\includegraphics[width=0.9\linewidth]{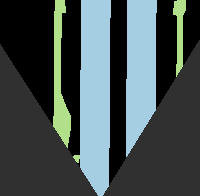}\linesep
\scriptsize{Ground Truth}\linesep
\end{minipage}
\hspace{-1.2ex}
    \caption{Qualitative segmentation results on the nuScenes benchmark. We visualize the class with the largest index $c$ which has occupancy probability $p_i > 0.5$. Black regions (outside field of view or no LiDAR returns) are ignored during evaluation. Please refer to the top legend of Fig. \ref{fig:intro} in the main paper for the meaning of different colors.}
    \label{fig:vis_results_nusc}
\end{figure*}

As shown in Fig. \ref{fig:vis_results_nusc}, we evaluate the segmentation performance of both dynamic and static objects on the nuScenes \cite{caesar2020nuscenes} benchmark. Previous approaches tend to produce relatively rough predictions and are more likely to be distorted when noises exist. For instance, cars which should be independent individuals are connected into a strip region. Compared with the existing methods, DiffBEV inherently has the power to reduce harmful noises, thus producing more accurate semantic maps and resolving fine-grained details such as the spatial separation between neighboring vehicles. 

\subsection{More Qualitative Analysis on the KITTI Datasets}
\begin{figure*}[htbp]
\centering
\begin{minipage}[c]{0.202\linewidth}
\centering
\includegraphics[height=0.53\linewidth]{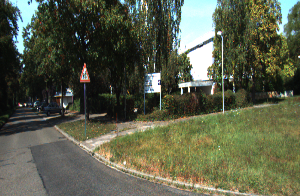}\linesep
\includegraphics[height=0.53\linewidth]{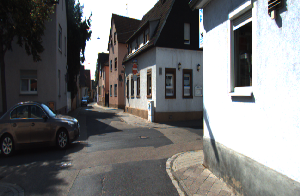}\linesep
\includegraphics[height=0.53\linewidth]{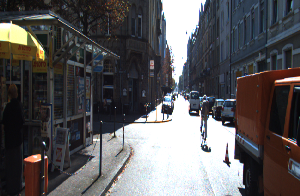}\linesep
\includegraphics[height=0.53\linewidth]{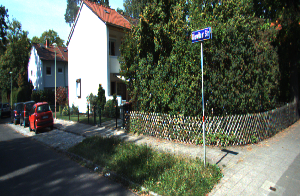}\linesep
\includegraphics[height=0.53\linewidth]{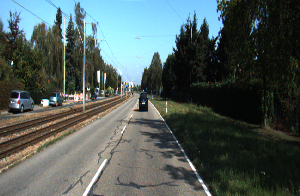}\linesep
\scriptsize{Image}\linesep
\end{minipage}
\hspace{-1.2ex}
\begin{minipage}[c]{0.121\linewidth}
\centering
\includegraphics[width=0.9\linewidth]{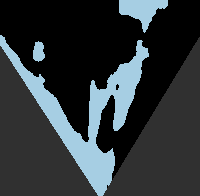}\linesep
\includegraphics[width=0.9\linewidth]{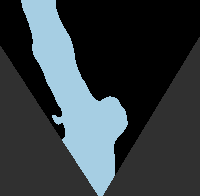}\linesep
\includegraphics[width=0.9\linewidth]{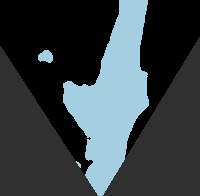}\linesep
\includegraphics[width=0.9\linewidth]{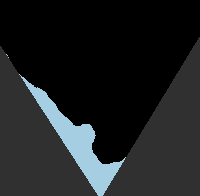}\linesep
\includegraphics[width=0.9\linewidth]{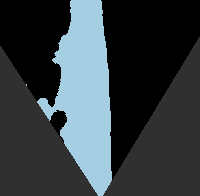}\linesep
\scriptsize{VPN\cite{pan2020cross}}\linesep
\end{minipage}
\hspace{-1.2ex}
\begin{minipage}[c]{0.121\linewidth}
\centering
\includegraphics[width=0.9\linewidth]{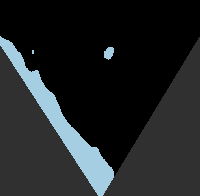}\linesep
\includegraphics[width=0.9\linewidth]{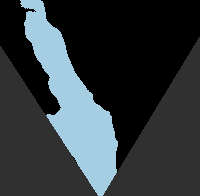}\linesep
\includegraphics[width=0.9\linewidth]{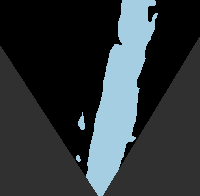}\linesep
\includegraphics[width=0.9\linewidth]{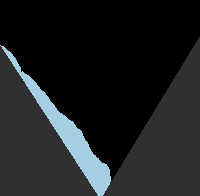}\linesep
\includegraphics[width=0.9\linewidth]{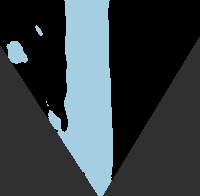}\linesep
\scriptsize{PYVA\cite{yang2021projecting}}\linesep
\end{minipage}
\hspace{-1.2ex}
\begin{minipage}[c]{0.121\linewidth}
\centering
\includegraphics[width=0.9\linewidth]{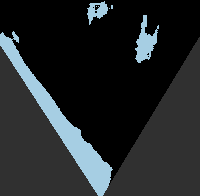}\linesep
\includegraphics[width=0.9\linewidth]{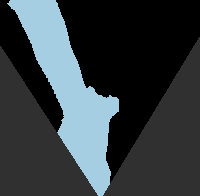}\linesep
\includegraphics[width=0.9\linewidth]{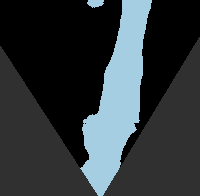}\linesep
\includegraphics[width=0.9\linewidth]{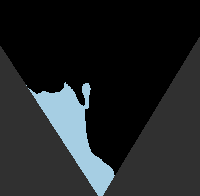}\linesep
\includegraphics[width=0.9\linewidth]{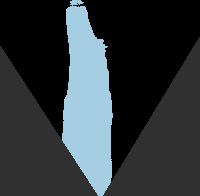}\linesep
\scriptsize{PON\cite{roddick2020predicting}}\linesep
\end{minipage}
\hspace{-1.2ex}
\begin{minipage}[c]{0.121\linewidth}
\centering
\includegraphics[width=0.9\linewidth]{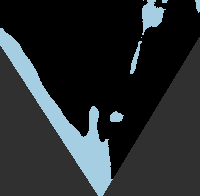}\linesep
\includegraphics[width=0.9\linewidth]{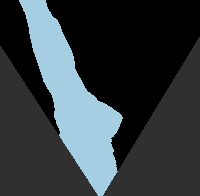}\linesep
\includegraphics[width=0.9\linewidth]{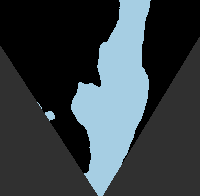}\linesep
\includegraphics[width=0.9\linewidth]{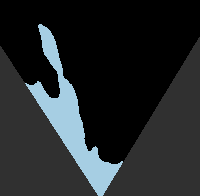}\linesep
\includegraphics[width=0.9\linewidth]{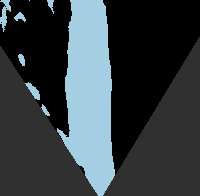}\linesep
\scriptsize{LSS\cite{philion2020lift}}\linesep
\end{minipage}
\hspace{-1.2ex}
\begin{minipage}[c]{0.121\linewidth}
\centering
\includegraphics[width=0.9\linewidth]{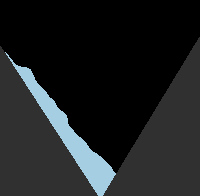}\linesep
\includegraphics[width=0.9\linewidth]{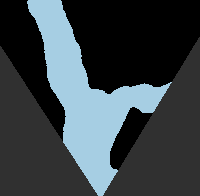}\linesep
\includegraphics[width=0.9\linewidth]{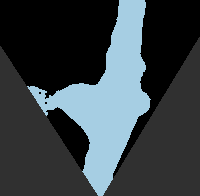}\linesep
\includegraphics[width=0.9\linewidth]{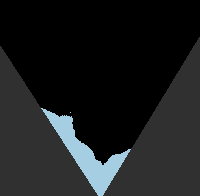}\linesep
\includegraphics[width=0.9\linewidth]{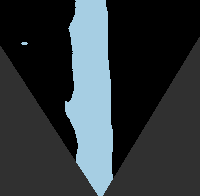}\linesep
\scriptsize{DiffBEV}\linesep
\end{minipage}
\hspace{-1.2ex}
\begin{minipage}[c]{0.121\linewidth}
\centering
\includegraphics[width=0.9\linewidth]{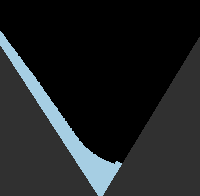}\linesep
\includegraphics[width=0.9\linewidth]{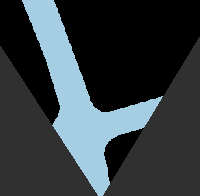}\linesep
\includegraphics[width=0.9\linewidth]{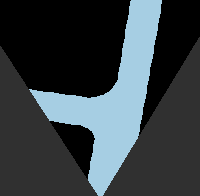}\linesep
\includegraphics[width=0.9\linewidth]{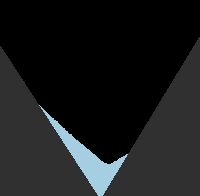}\linesep
\includegraphics[width=0.9\linewidth]{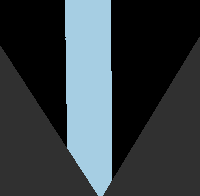}\linesep
\scriptsize{Ground Truth}\linesep
\end{minipage}
\hspace{-1.2ex}
    \caption{Qualitative results on the KITTI Raw benchmark. We evaluate the performance of the static road layout estimation.}
    \label{fig:vis_results_raw}
\end{figure*}

\begin{figure*}[htbp]
\centering
\begin{minipage}[c]{0.206\linewidth}
\centering
\includegraphics[height=0.53\linewidth]{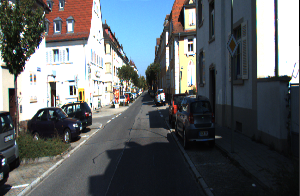}\linesep
\includegraphics[height=0.53\linewidth]{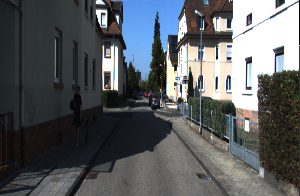}\linesep
\includegraphics[height=0.53\linewidth]{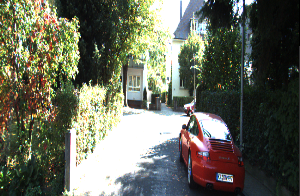}\linesep
\includegraphics[height=0.53\linewidth]{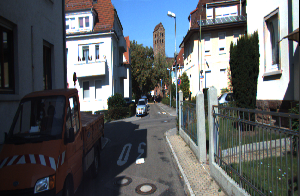}\linesep
\includegraphics[height=0.53\linewidth]{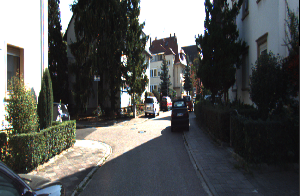}\linesep
\scriptsize{Image}\linesep
\end{minipage}
\hspace{-1.2ex}
\begin{minipage}[c]{0.121\linewidth}
\centering
\includegraphics[height=0.9\linewidth]{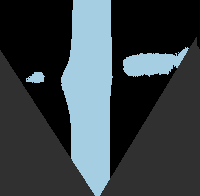}\linesep
\includegraphics[height=0.9\linewidth]{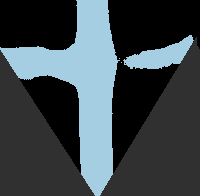}\linesep
\includegraphics[height=0.9\linewidth]{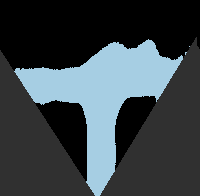}\linesep
\includegraphics[height=0.9\linewidth]{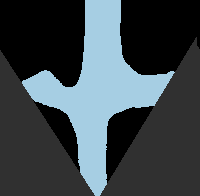}\linesep
\includegraphics[height=0.9\linewidth]{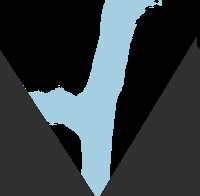}\linesep
\scriptsize{VPN\cite{pan2020cross}}\linesep
\end{minipage}
\hspace{-1.2ex}
\begin{minipage}[c]{0.121\linewidth}
\centering
\includegraphics[height=0.9\linewidth]{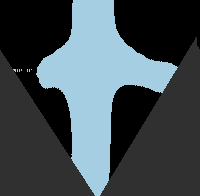}\linesep
\includegraphics[height=0.9\linewidth]{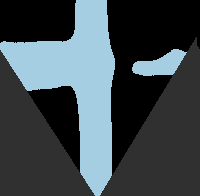}\linesep
\includegraphics[height=0.9\linewidth]{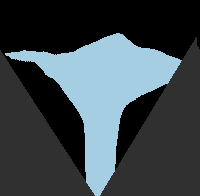}\linesep
\includegraphics[height=0.9\linewidth]{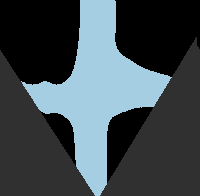}\linesep
\includegraphics[height=0.9\linewidth]{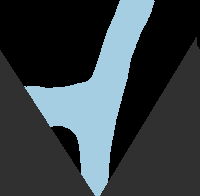}\linesep
\scriptsize{PYVA\cite{yang2021projecting}}\linesep
\end{minipage}
\hspace{-1.2ex}
\begin{minipage}[c]{0.121\linewidth}
\centering
\includegraphics[height=0.9\linewidth]{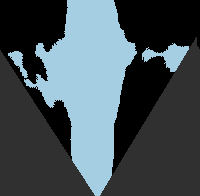}\linesep
\includegraphics[height=0.9\linewidth]{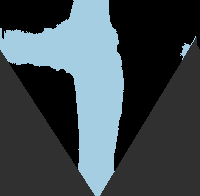}\linesep
\includegraphics[height=0.9\linewidth]{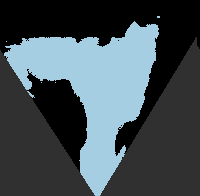}\linesep
\includegraphics[height=0.9\linewidth]{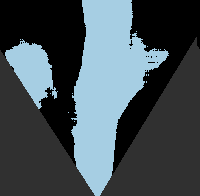}\linesep
\includegraphics[height=0.9\linewidth]{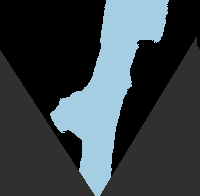}\linesep
\scriptsize{PON\cite{roddick2020predicting}}\linesep
\end{minipage}
\hspace{-1.2ex}
\begin{minipage}[c]{0.121\linewidth}
\centering
\includegraphics[height=0.9\linewidth]{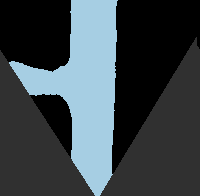}\linesep
\includegraphics[height=0.9\linewidth]{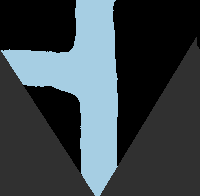}\linesep
\includegraphics[height=0.9\linewidth]{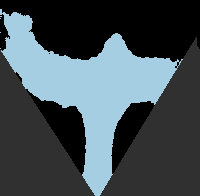}\linesep
\includegraphics[height=0.9\linewidth]{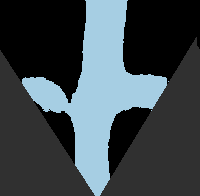}\linesep
\includegraphics[height=0.9\linewidth]{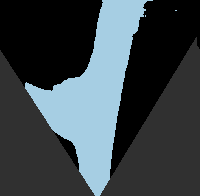}\linesep
\scriptsize{LSS\cite{philion2020lift}}\linesep
\end{minipage}
\hspace{-1.2ex}
\begin{minipage}[c]{0.121\linewidth}
\centering
\includegraphics[height=0.9\linewidth]{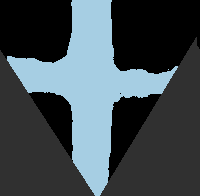}\linesep
\includegraphics[height=0.9\linewidth]{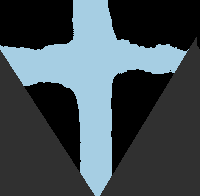}\linesep
\includegraphics[height=0.9\linewidth]{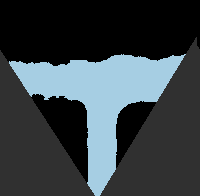}\linesep
\includegraphics[height=0.9\linewidth]{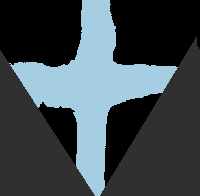}\linesep
\includegraphics[height=0.9\linewidth]{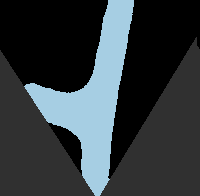}\linesep
\scriptsize{DiffBEV}\linesep
\end{minipage}
\hspace{-1.2ex}
\begin{minipage}[c]{0.121\linewidth}
\centering
\includegraphics[height=0.9\linewidth]{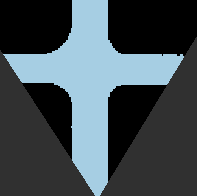}\linesep
\includegraphics[height=0.9\linewidth]{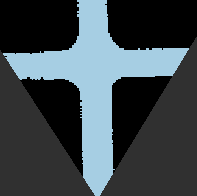}\linesep
\includegraphics[height=0.9\linewidth]{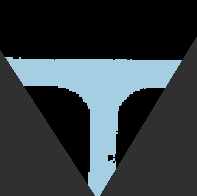}\linesep
\includegraphics[height=0.9\linewidth]{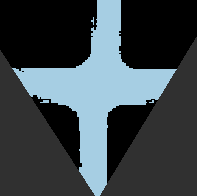}\linesep
\includegraphics[height=0.9\linewidth]{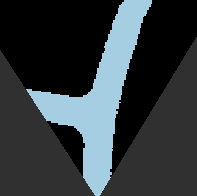}\linesep
\scriptsize{Ground Truth}\linesep
\end{minipage}
\hspace{-1.2ex}
    \caption{Qualitative results on the KITTI Odometry benchmark. We evaluate the performance of the static road layout estimation.}
    \label{fig:vis_results_odometry}
\end{figure*}

\begin{figure*}[htbp]
\centering
\begin{minipage}[c]{0.202\linewidth}
\centering
\includegraphics[height=0.53\linewidth]{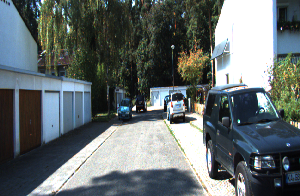}\linesep
\includegraphics[height=0.53\linewidth]{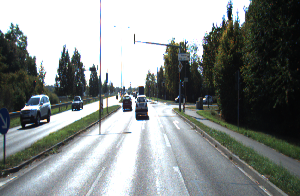}\linesep
\includegraphics[height=0.53\linewidth]{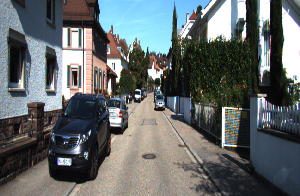}\linesep
\includegraphics[height=0.53\linewidth]{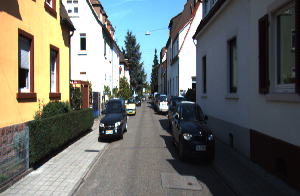}\linesep
\includegraphics[height=0.53\linewidth]{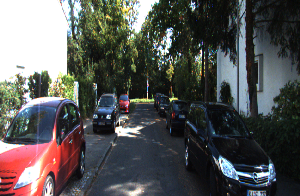}\linesep
\scriptsize{Image}\linesep
\end{minipage}
\hspace{-1.2ex}
\begin{minipage}[c]{0.121\linewidth}
\centering
\includegraphics[width=0.9\linewidth]{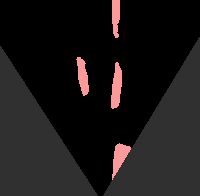}\linesep
\includegraphics[width=0.9\linewidth]{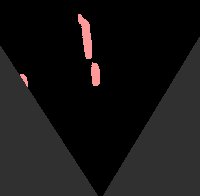}\linesep
\includegraphics[width=0.9\linewidth]{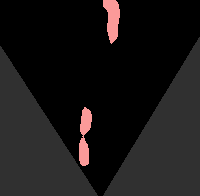}\linesep
\includegraphics[width=0.9\linewidth]{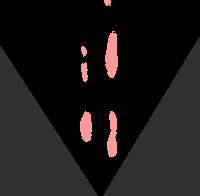}\linesep
\includegraphics[width=0.9\linewidth]{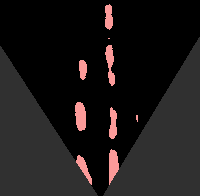}\linesep
\scriptsize{VPN\cite{pan2020cross}}\linesep
\end{minipage}
\hspace{-1.2ex}
\begin{minipage}[c]{0.121\linewidth}
\centering
\includegraphics[width=0.9\linewidth]{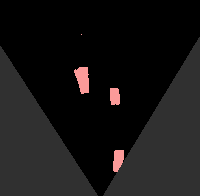}\linesep
\includegraphics[width=0.9\linewidth]{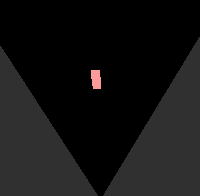}\linesep
\includegraphics[width=0.9\linewidth]{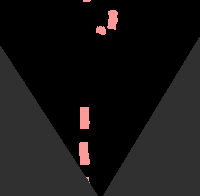}\linesep
\includegraphics[width=0.9\linewidth]{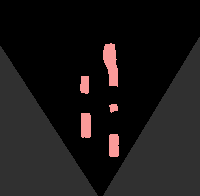}\linesep
\includegraphics[width=0.9\linewidth]{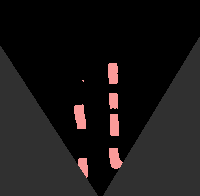}\linesep
\scriptsize{PYVA\cite{yang2021projecting}}\linesep
\end{minipage}
\hspace{-1.2ex}
\begin{minipage}[c]{0.121\linewidth}
\centering
\includegraphics[width=0.9\linewidth]{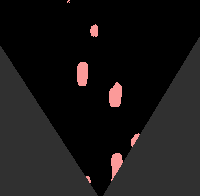}\linesep
\includegraphics[width=0.9\linewidth]{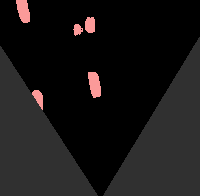}\linesep
\includegraphics[width=0.9\linewidth]{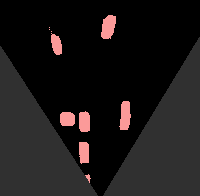}\linesep
\includegraphics[width=0.9\linewidth]{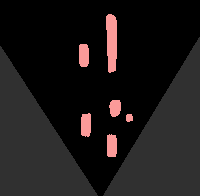}\linesep
\includegraphics[width=0.9\linewidth]{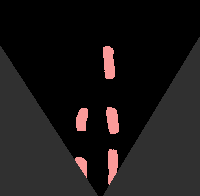}\linesep
\scriptsize{PON\cite{roddick2020predicting}}\linesep
\end{minipage}
\hspace{-1.2ex}
\begin{minipage}[c]{0.121\linewidth}
\centering
\includegraphics[width=0.9\linewidth]{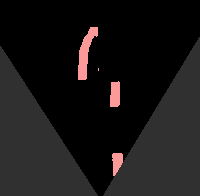}\linesep
\includegraphics[width=0.9\linewidth]{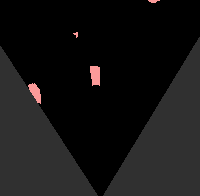}\linesep
\includegraphics[width=0.9\linewidth]{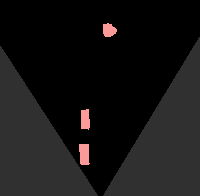}\linesep
\includegraphics[width=0.9\linewidth]{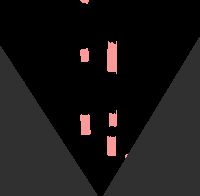}\linesep
\includegraphics[width=0.9\linewidth]{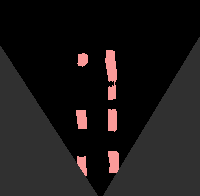}\linesep
\scriptsize{LSS\cite{philion2020lift}}\linesep
\end{minipage}
\hspace{-1.2ex}
\begin{minipage}[c]{0.121\linewidth}
\centering
\includegraphics[width=0.9\linewidth]{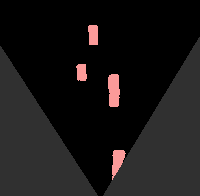}\linesep
\includegraphics[width=0.9\linewidth]{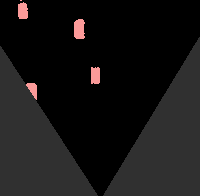}\linesep
\includegraphics[width=0.9\linewidth]{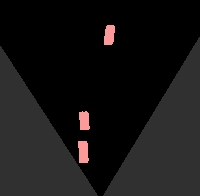}\linesep
\includegraphics[width=0.9\linewidth]{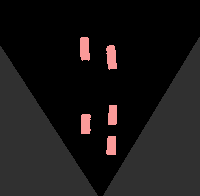}\linesep
\includegraphics[width=0.9\linewidth]{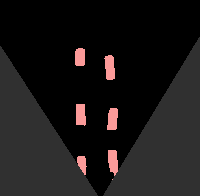}\linesep
\scriptsize{DiffBEV}\linesep
\end{minipage}
\hspace{-1.2ex}
\begin{minipage}[c]{0.121\linewidth}
\centering
\includegraphics[width=0.9\linewidth]{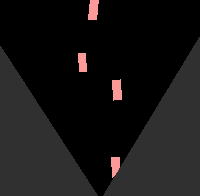}\linesep
\includegraphics[width=0.9\linewidth]{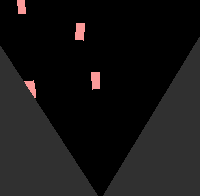}\linesep
\includegraphics[width=0.9\linewidth]{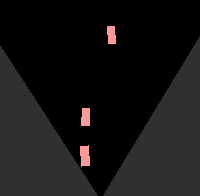}\linesep
\includegraphics[width=0.9\linewidth]{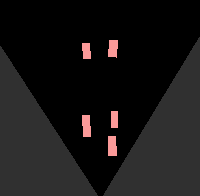}\linesep
\includegraphics[width=0.9\linewidth]{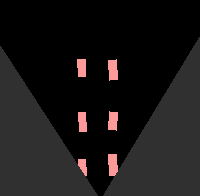}\linesep
\scriptsize{Ground Truth}\linesep
\end{minipage}
\hspace{-1.2ex}
    \caption{Qualitative results on the KITTI 3D Object benchmark. We evaluate the segmentation performance on dynamic vehicles. Compared with the previous methods, DiffBEV reduces noises better and presents semantic maps for dynamic vehicles with clearer edges and more accurate geometric positions.}
    \label{fig:vis_results_object}
\end{figure*}

As indicated in Fig. \ref{fig:vis_results_raw} and Fig. \ref{fig:vis_results_odometry}, we evaluate the performance of static road estimation on the KITTI Raw \cite{geiger2012we} and KITTI Odometry \cite{behley2019semantickitti} datasets, respectively. Fig. \ref{fig:vis_results_object} shows the segmentation performance on dynamic vehicles on the KITTI 3D Object \cite{geiger2012we} benchmark. For static roads and dynamic vehicles, DiffBEV outperforms the previous methods and provides semantic maps with clearer edges and more accurate positions, which confirms that DiffBEV is able to effectively learn the structural information of objects.

\end{document}